%% file: main.tex
\documentclass[10pt,twocolumn,letterpaper]{article}

\usepackage{iccv}
\usepackage{times}
\usepackage{epsfig}
\usepackage{graphicx}
\usepackage{amsmath}
\usepackage{amssymb}
\usepackage{enumitem}
\usepackage{boldline}
\usepackage{subfig}
\usepackage{multirow}
\usepackage{xcolor}
\usepackage{pifont}
\usepackage{placeins}
\usepackage{enumitem}
\usepackage[linewidth=1pt]{mdframed}
\usepackage[title]{appendix}
\usepackage{booktabs}

\usepackage[square,numbers,sort&compress]{natbib}

\usepackage[pagebackref=true,breaklinks=true,letterpaper=true,colorlinks,bookmarks=false,citecolor=blue]{hyperref}

\newcommand{\cmark}{\ding{51}}%
\newcommand{\xmark}{\ding{55}}%

\newcommand{\OurDataset}{\text{COCO-BISON}\xspace}

\newcommand{\GCSI}{\text{BISON}\xspace}

\iccvfinalcopy 

\def\iccvPaperID{***} 

\ificcvfinal\fi
\begin{document}

\title{Evaluating Text-to-Image Matching using Binary Image Selection (BISON)}

\author{Hexiang Hu\thanks{This work was performed while Hexiang Hu was at Facebook.}\\
	University of Southern California\\
	{\tt\small hexiangh@usc.edu}
	\and
	Ishan Misra\\
	Facebook AI Research\\
	{\tt\small imisra@fb.com}
	\and
	Laurens van der Maaten\\
	Facebook AI Research\\
	{\tt\small lvdmaaten@fb.com}
}

\maketitle

\begin{abstract}
Providing systems the ability to relate linguistic and visual content is one of the hallmarks of computer vision. Tasks such as text-based image retrieval and image captioning were designed to test this ability, but come with evaluation measures that have high variance or are difficult to interpret. We study an alternative task for systems that match text and images: given a text query, the system is asked to select the image that best matches the query from a pair of semantically similar images. The system's accuracy on this Binary Image SelectiON (BISON) task is interpretable, eliminates the reliability problems of retrieval evaluations, and focuses on the system's ability to understand fine-grained visual structure. We gather a BISON dataset that complements the COCO dataset and use it to evaluate modern text-based image retrieval and image captioning systems. Our results provide novel insights into the performance of these systems.
\end{abstract}

\input{intro}

\input{captioning_analysis}
\input{dataset}
\input{system_benchmarking}
\input{related}
\input{bison_analysis}
\input{conclusion}

\section*{Acknowledgements}
We thank Devi Parikh, Marcus Rohrbach, Brian Knott, and anonymous reviewers for comments on early versions of this paper.

{\small
	\bibliographystyle{ieee}
	\bibliography{references}
}

\appendix
\clearpage

\include{supp_content}

\end{document}

%% file: intro.tex

\section{Introduction}

Understanding the relation between linguistic and visual content is a fundamental goal of computer vision. This goal has motivated a large body of research focusing on tasks such as text-based image retrieval~\cite{hodosh2013framing,li2017ngrams} and image captioning~\cite{hodosh2013framing,karpathy2015deep,vinyals2015show,Xu2015ShowAA}. Both these tasks have challenges in terms of evaluation: in particular, the open-ended nature of image captioning tasks makes it difficult to develop evaluation measures~\cite{anderson2016spice,vedantam2015cider} that reliably and accurately measure image-text relevance without also considering other abilities, such as fluency in language generation~\cite{lsun-coco-workshop}. Text-based image retrieval does not have these problems, but is unreliable because retrieval datasets are only partially labeled: they incorrectly assume that images that are not positively labeled for a given text query are negative examples. Our analysis of the errors of caption-based image retrieval systems reveals that more than half of these ``errors'' are due to errors in the evaluation, that is, due to ``negative'' images being accurately described by the text query; see Section~\ref{sec:captioning_analysis} for a detailed analysis.

\begin{figure}[!t]
\centering
\includegraphics[width=\linewidth]{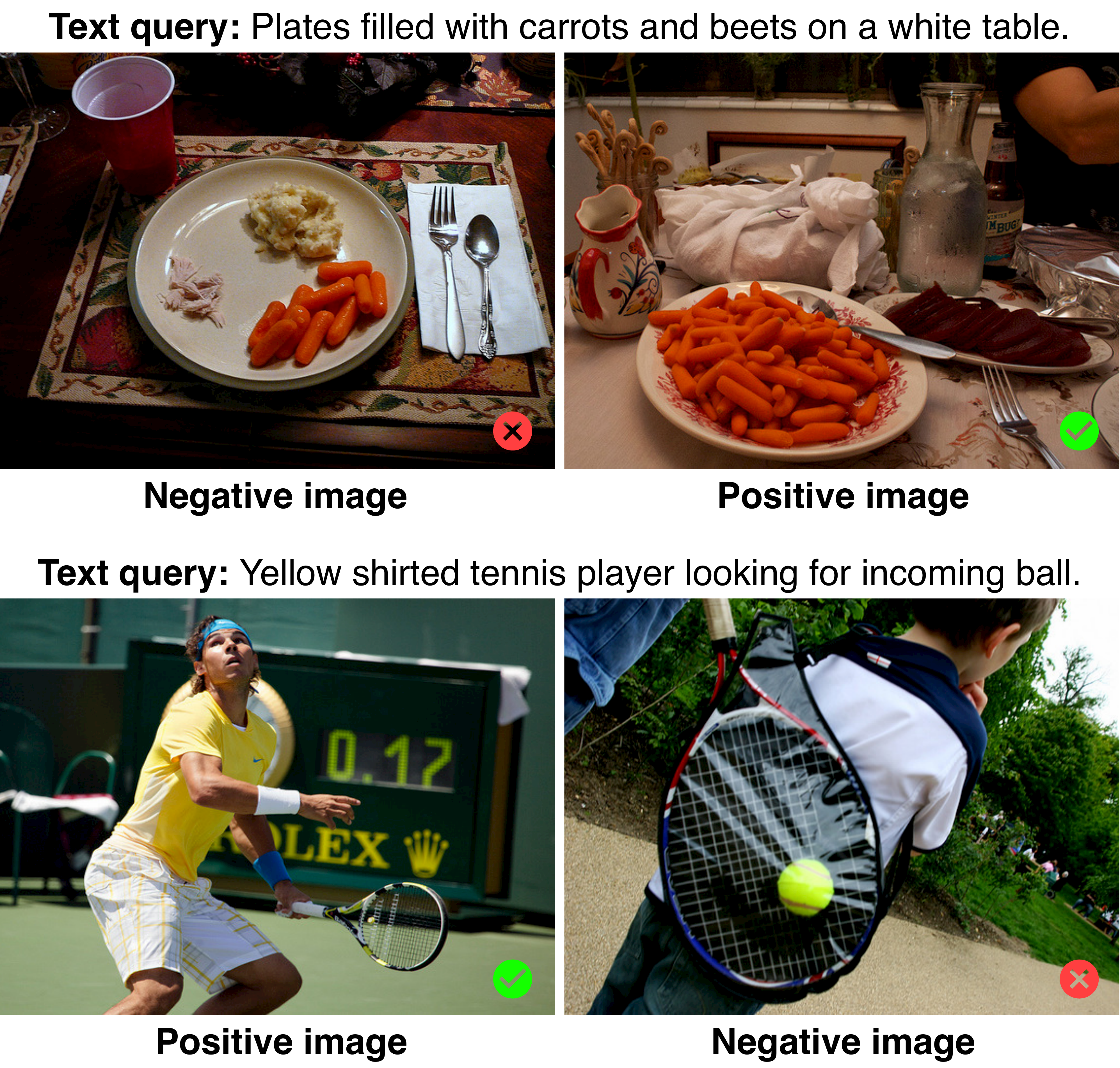}
\caption{\textbf{Binary Image SelectiON (BISON):} Given a \emph{text query}, the system must select which of two images best matches the caption. This task evaluates fine-grained visual grounding. The BISON accuracy of a system is the proportion of examples for which the system correctly chooses the \emph{positive image} ({\color{green}\cmark}) over the \emph{negative image} ({\color{red}\xmark}).}
\label{fig:teaser}
\end{figure}

Motivated by this issue, we propose an alternative task to evaluate systems that match textual and visual content, called \emph{Binary Image SelectiON (BISON)}. In BISON, the system is provided with two similar images and a fine-grained text query that describes one image but not the other. The system needs to select which of the two images is described in the text query; see Figure~\ref{fig:teaser}. The performance of the system is measured in terms of its binary classification accuracy of selecting the correct image. Indeed, BISON can be viewed as a variant of text-based image retrieval in which positive and negative examples are explicitly labeled. BISON can be used as an auxiliary evaluation of generative (captioning) and discriminative (retrieval) vision-language models, which facilitates its use in conjunction with existing evaluations. BISON accuracy differs from existing tasks in that it is reliable, easy to interpret, and focuses on fine-grained visual content.

To facilitate binary image selection experiments, we collected the \emph{\OurDataset dataset} using the images and captions in the existing COCO~\cite{chen2015microsoft} validation set. By using both the text and images from the COCO dataset, we ensure that \OurDataset has a similar distribution as COCO --- this allows for the evaluation of COCO-trained models on the \OurDataset dataset. We use the \OurDataset dataset to evaluate state-of-the-art text-based image retrieval and image captioning systems, shedding new light on the performance of these systems. For example, in contrast to prior work, we find these systems are not as good as humans in matching visual and (detailed) linguistic content.


%% file: captioning_analysis.tex

\begin{table}[t]
	\centering
	\label{tab:retrieval_error_analysis}
	\begin{tabular}{cccc}\toprule
		\bf Recall@1 & \bf Human & \bf Number & \bf Percentage \\
		\midrule
		Incorrect & Incorrect & 165 & 43.9\% \\
		Incorrect & Correct & 211 & 56.1\% \\\bottomrule
	\end{tabular}
	\caption{\textbf{Analysis of the recall@1 text-based image retrieval measure.} We run SCAN t2i~\cite{lee2018stacked} image retrieval on the COCO captions~\cite{chen2015microsoft} validation set and ask humans to analyze all $376$ retrieval ``errors'' according to the recall@1 retrieval measure. Human annotators mark $56.1\%$ of these ``errors'' as correct retrievals.}
	\label{tab:retrieval_error_analysis}
\end{table}  

\begin{figure}[t]
	\centering
	\includegraphics[width=\linewidth]{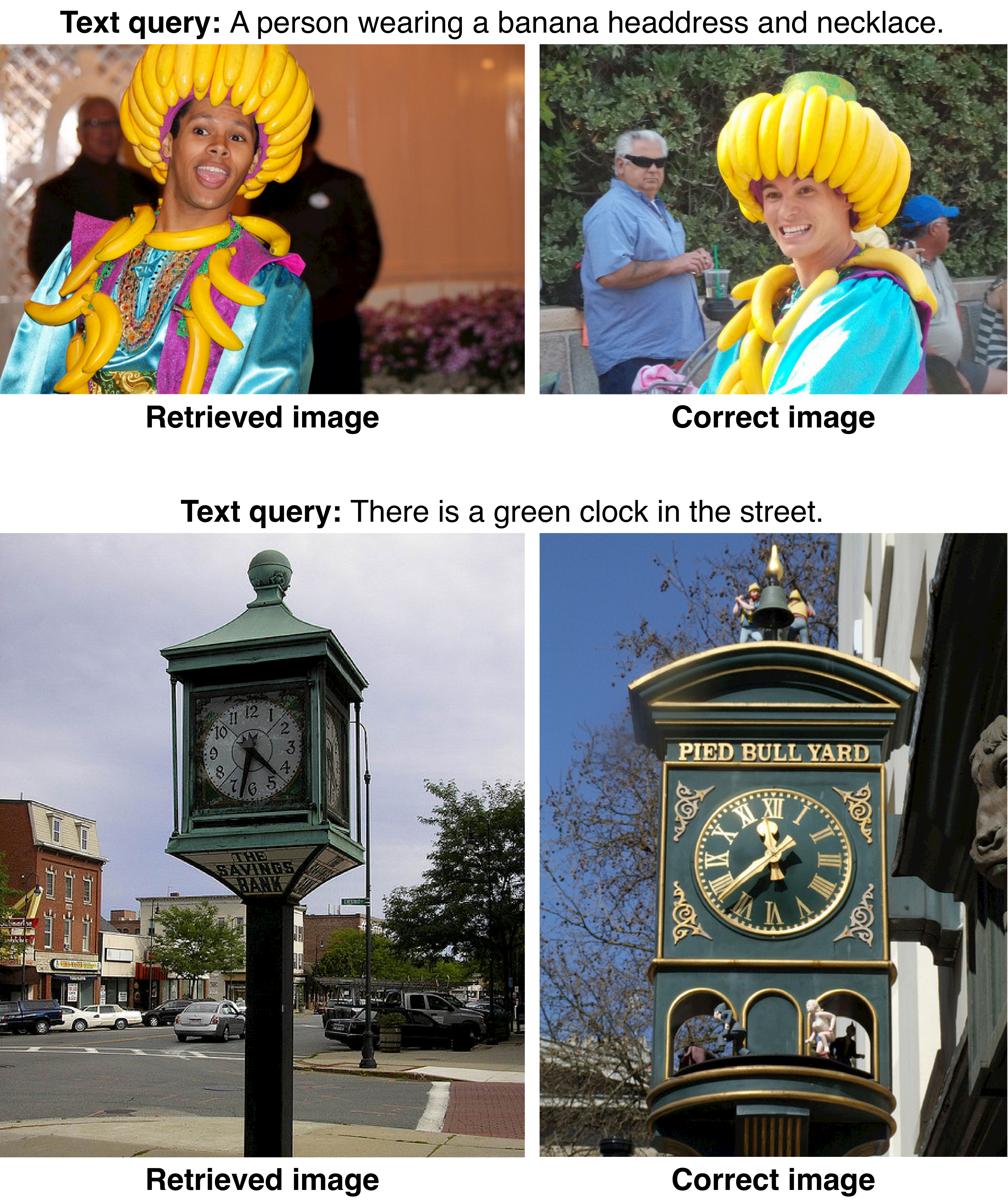}
	\caption{\textbf{Examples of ``incorrect'' image retrievals given a text query:} Image retrieved from COCO captions validation set by SCAN t2i~\cite{lee2018stacked} (left) given a text query (top), and the image that should have been retrieved for that query (right) according to the recall@1 retrieval measure. The examples show that in retrieval evaluations, correctly retrieved images may be counted as incorrect because the retrieval task erroneously assumes that all but one of the images are negative examples for the text query.} 
	\label{fig:retrieval_negatives}
\end{figure}

\begin{figure*}[t]
	\centering
	\includegraphics[width=0.95\linewidth]{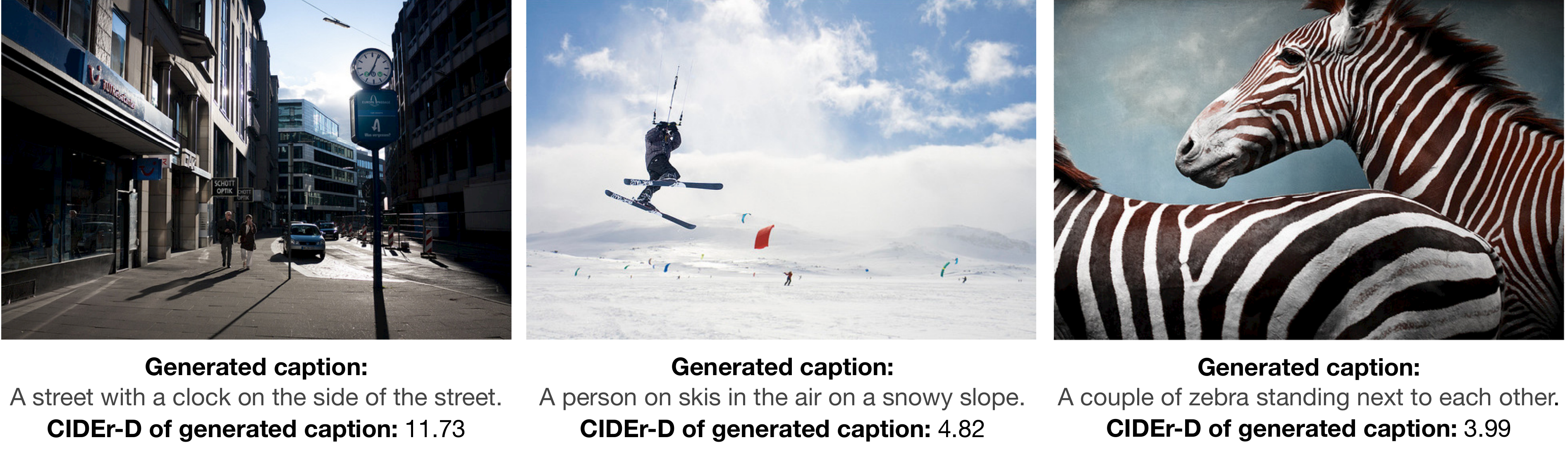}
	\caption{\textbf{CIDEr-D score of correctly generated captions:} All pairs have a correctness score of 4.0 (per human annotators) but a low CIDEr-D score because the generated caption does not match the reference captions in the COCO dataset.}
	\label{fig:correctness_of_cider}
\end{figure*}

\begin{figure}[t]
\begin{minipage}[t]{\linewidth}
	\centering
	\subfloat{\includegraphics[width=0.5\linewidth]{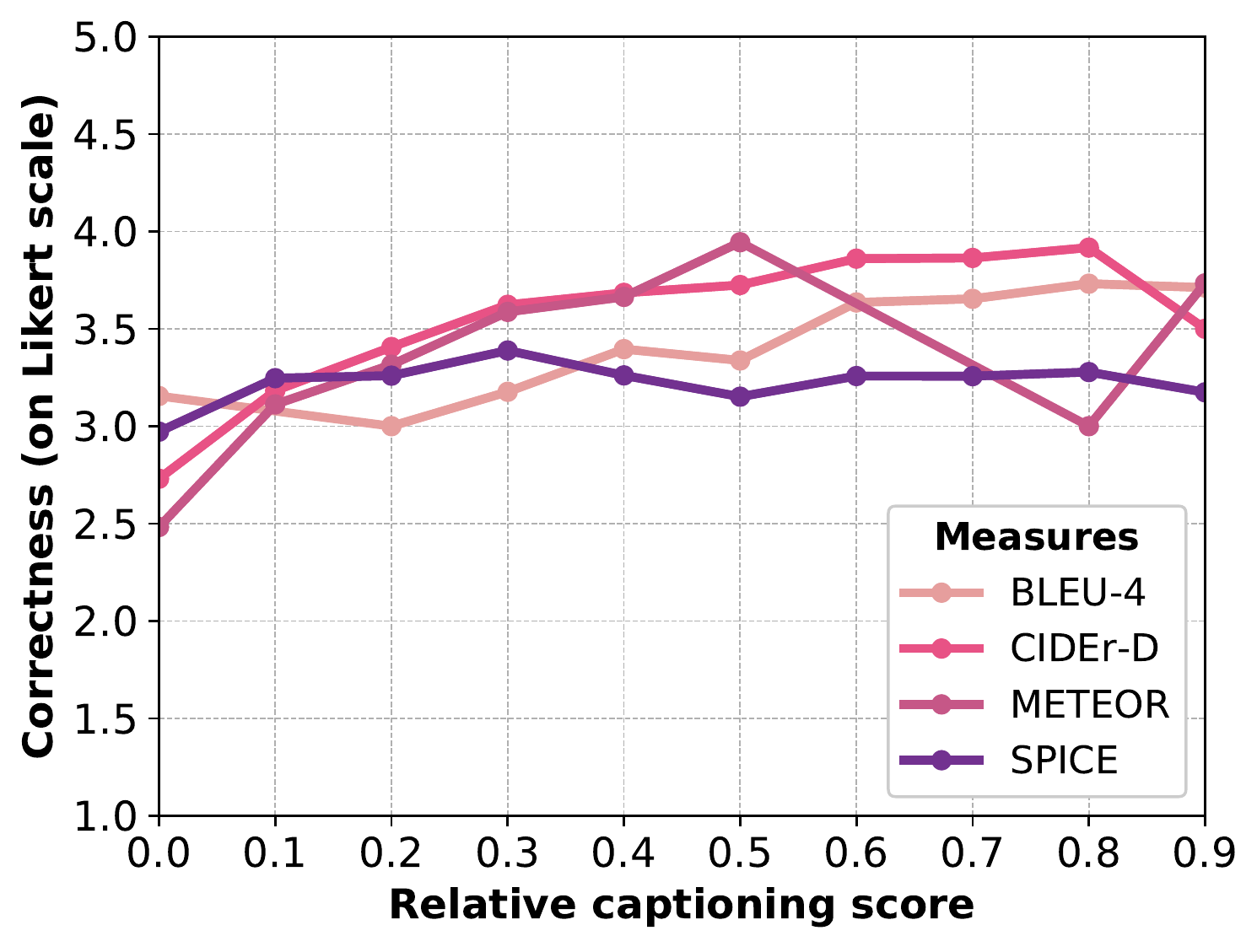}}
	\subfloat{\includegraphics[width=0.5\linewidth]{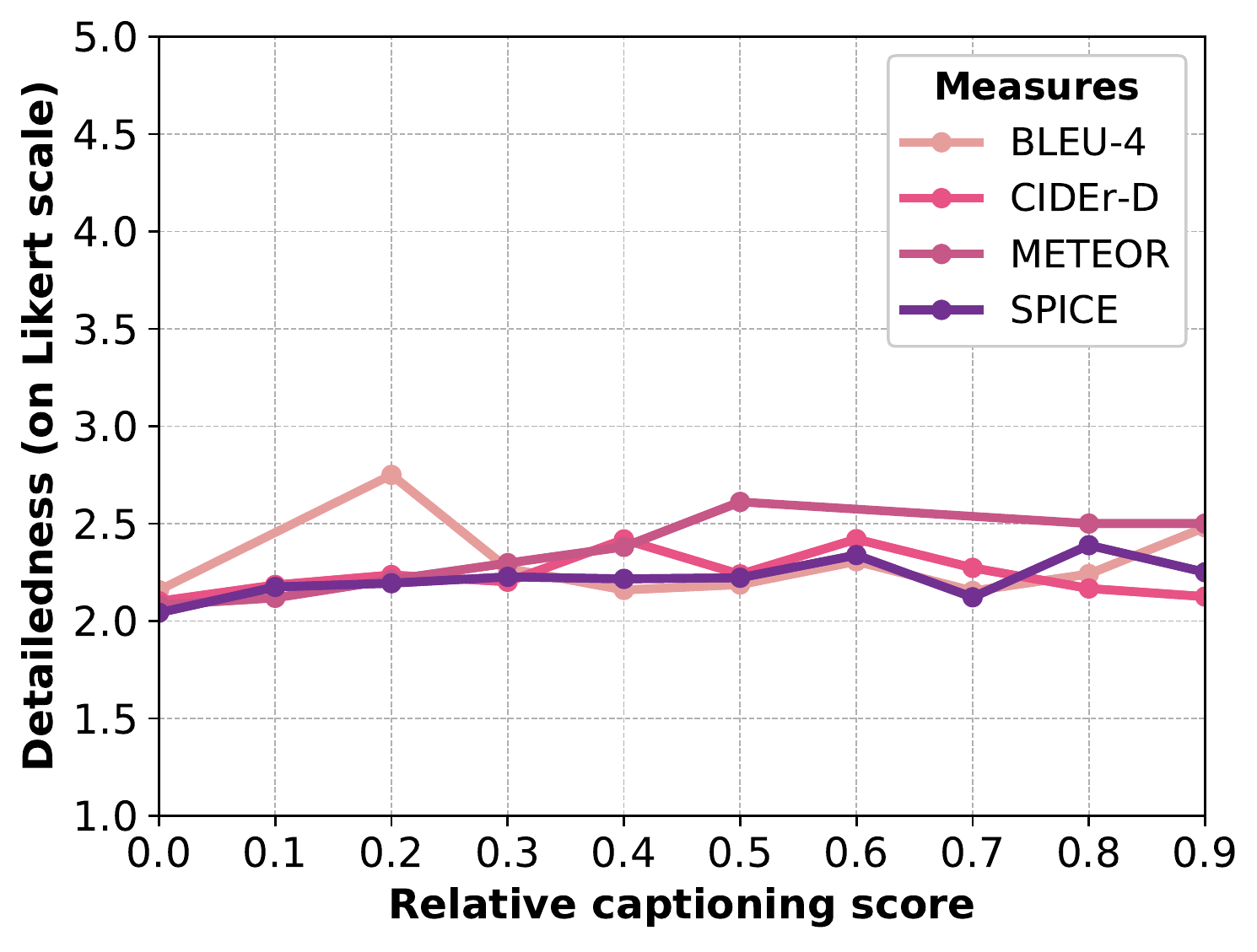}}
	\caption{\textbf{Correctness} (left) \textbf{and detailedness} (right) \textbf{of generated captions as a function of their captioning scores.} Captions were generated using the UpDown~\cite{Anderson2017BottomUpAT} captioning system. Correctness and detailedness of the generated captions were rated on a Likert scale (from 1 to 5) by human annotators. The average correctness and detailedness scores are 3.266 and 2.203, respectively.}
	\label{fig:caption_human_eval}
\end{minipage}
\end{figure}

\section{Analyzing Retrieval and Captioning Tasks}
\label{sec:captioning_analysis}
We performed two experiments to identify the limitations of popular evaluations of vision-and-language systems via text-based image retrieval and image captioning.

\par \noindent \textbf{Text-based image retrieval.} Evaluations via text-based image retrieval use a single positive image for each text query and assume all other images in the dataset are negative examples for that query~\cite{hodosh2013framing,li2017ngrams}. This assumption is often incorrect, in particular, when the image datasets is large. To assess the severity this problem, we performed an experiment in which we analyzed the ``errors'' made by the state-of-the-art SCAN t2i retrieval system~\cite{lee2018stacked} on the COCO captions validation set. We presented each incorrectly retrieved image along with the text query to a set of human annotators, asking them to indicate if the text query appropriately describes the content of the image. The results of this experiment are presented in Table~\ref{tab:retrieval_error_analysis} and suggest that more than half of the ``errors'' made by the SCAN t2i system are, in fact, not errors: the retrieved images are erroneously marked as incorrect due to the lack of explicit negative annotations. Figure~\ref{fig:retrieval_negatives} illustrates the problem by showing two examples of a text query, an ``incorrectly'' retrieved image, and the image labeled as positive for the query. Our results suggest that image retrieval measures are very unreliable.

\par \noindent \textbf{Image captioning.} Captioning evaluation measures such as BLEU-4~\cite{papineni2002bleu}, CIDEr-D~\cite{vedantam2015cider}, METEOR~\cite{banerjee2005meteor}, and SPICE~\cite{anderson2016spice} compare a generated caption to a collection of reference captions. As a result, the evaluations may be sensitive to changes in the reference caption set and incorrectly assess the semantics of the generated caption. We perform an analysis designed to study these effects on the COCO captions validation set by asking human annotators to assess image captions generated by the state-of-the-art UpDown~\cite{Anderson2017BottomUpAT,shuster2018engaging} captioning system\footnote{We performed the same experiment using other captioning methods. The results of these experiments were qualitatively similar, and are presented in the supplemental material.}. Specifically, we followed the COCO guidelines for human evaluation~\cite{lsun-coco-workshop} and asked annotators to evaluate the ``correctness'' of image-caption pairs on a Likert scale from 1 (low) to 5 (high). We asked a second set of annotators to evaluate the ``detailedness'' of captions (without showing them the image) on the same Likert scale.



Figure~\ref{fig:caption_human_eval} shows the resulting correctness and detailedness assessments as a function of four captioning scores (BLEU-4, CIDEr-D, METEOR, and SPICE) that were normalized to lie between $0$ and $1$. The results in the figure suggest that captioning scores do not correlate with the correctness of generated captions very well, and do not encourage generated captions to provide a detailed description of the image. Figure~\ref{fig:correctness_of_cider} shows three examples of generated captions that have a low CIDEr-D score even though they are correct according to human annotators. The examples highlight the limitations of using a handful of reference captions to evaluate captioning systems: the reference captions do not capture all visual content and all the different ways in which that content can be described~\cite{berg2012understanding,misra2016seeing}. This leads captioning measures to reward systems for generating generic captions.



%% file: dataset.tex
\begin{figure*}[t]
	\centering
	\includegraphics[width=\linewidth]{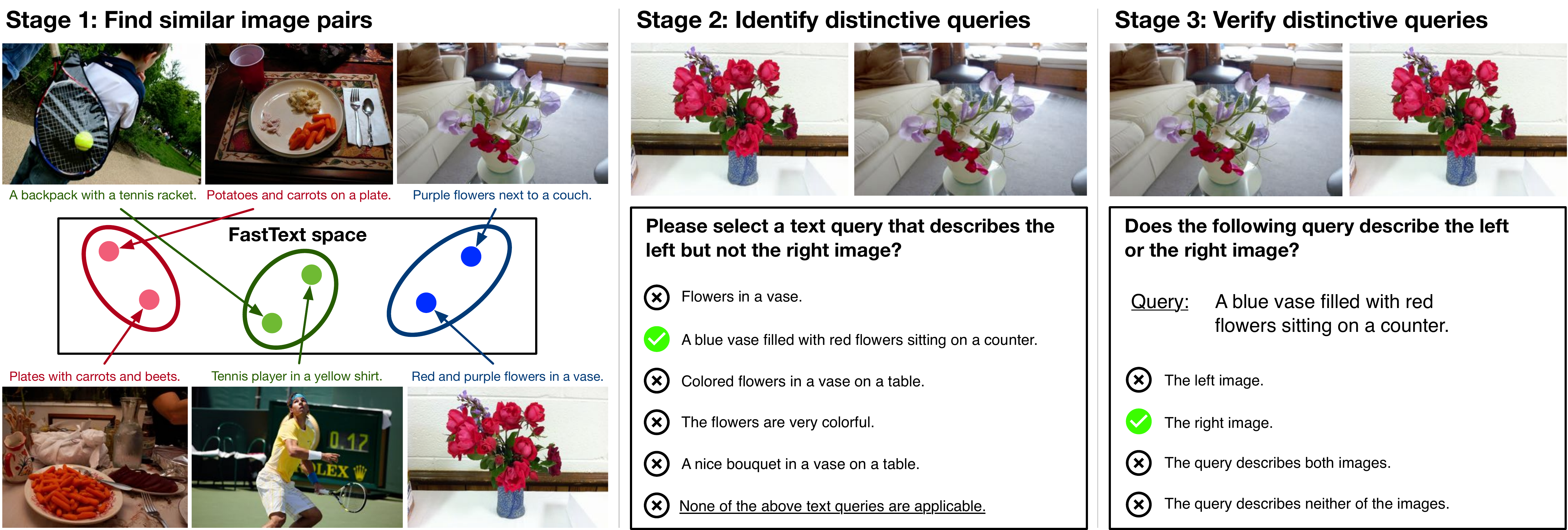}
	\caption{\textbf{Illustration of COCO-BISON dataset collection:} We collect annotations for our binary image selection task on top of the COCO Captions dataset. We first find pairs of semantically similar images based on the similarities between their reference captions. Annotators then select a text query that describes only one of the images in a pair. Finally, we validate the annotations by asking annotators to select the correct image given the text query. See Section~\ref{sec:dataset_collection} for details.}
	\label{fig:overview_plot}
\end{figure*}

\section{The COCO-BISON Dataset}
\label{sec:dataset_collection}
The goal of binary image selection (BISON) is to provide a reliable and interpretable evaluation task for systems that relate images and text, with a focus on fine-grained visual content. To this end, following~\cite{goyal2017making}, we collected BISON annotations for the validation split of the COCO captions dataset~\cite{chen2015microsoft}. 

\subsection{Collection of BISON Annotations}
\label{ref:collecting_dataset}
Figure~\ref{fig:overview_plot} illustrates the three main stages of our pipeline for collecting binary image selection annotations.

\noindent\textbf{1. Collect pairs of semantically similar images.} We construct a semantic representation for each image in the COCO validation set by averaging word embeddings (obtained using FastText~\cite{joulin2016fasttext}) of all the words in all captions associated with the image. We use these representations to find the semantically most similar image for each image in the dataset via nearest neighbor search. We label the query image as positive and its nearest neighbor as negative.

\noindent\textbf{2. Identify text queries that distinguish positive and negative images.} We present human annotators with an interface\footnote{Screenshots of the annotation interface in the supplementary material.} that shows: (1) a positive image, (2) the corresponding negative image, and (3) the five captions associated with the positive image in the COCO captions dataset. We ask the annotators to select a text query from the set of five captions that describes the positive image \emph{but not} the negative image, or to select ``none of the above'' if no discriminative text query exists. Unless annotators select the latter option, each of their annotations produces a query-positive-negative triple. We discard all image pairs for which annotators indicated no discriminative text query exists.

\noindent\textbf{3. Verify correctness of the query-positive-negative triples.} 
To ensure the validity of each query-positive-negative triple, we presented a different set of human annotators with trials that contained the positive and negative images and the query selected in stage 2. We asked the annotators whether the text query describes\footnote{In the verification stage, the annotators do not know which image is positive and which one is negative.}: (1) the positive image, (2) the negative image, (3) both images, or (4) neither of the images. Each verification trial was performed by two annotators; we only accepted the corresponding BISON example if both annotators correctly selected the positive image given the text query.

The query-positive-negative triples thus collected form binary image selection (BISON) examples, two of which are shown in Figure~\ref{fig:teaser}. The COCO-BISON dataset and corresponding evaluation code is publicly available from \url{http://hexianghu.com/bison/}.

\subsection{Dataset Characteristics}
\label{ref:dataset_stats}


Table~\ref{tab_stats} presents key statistics of our \OurDataset dataset; for reference, it also shows the statistics of the validation splits of two popular captioning datasets. As shown in the table, our three-stage annotation procedure produced a BISON example for $38,680$ of the $ 40,504$ the images in the COCO captions validation set ($\approx 95.5\%$).

\begin{table}[t]
	\centering
	\small
	\setlength{\tabcolsep}{2pt}
	\begin{tabular}{lccc}\toprule
		& \bf Flickr-30K & \bf COCO val & \bf\OurDataset \\\midrule
		Number of examples & 5,070 & 202,654 &  54,253 \\
		Unique images & 1,014 & 40,504 & 38,680 \\
		Unique captions & 5,068 & 197,792 & 45,218 \\\bottomrule
	\end{tabular}
	\caption{\textbf{Key statistics of the COCO-BISON dataset:} The statistics of the Flickr-30K~\cite{young2014image} and COCO Captions~\cite{chen2015microsoft} validation sets are shown for reference.}
	\label{tab_stats}
\end{table}

\begin{figure}[t]
	\centering
	\vspace{-1em}
	\subfloat{\includegraphics[width=0.51\linewidth]{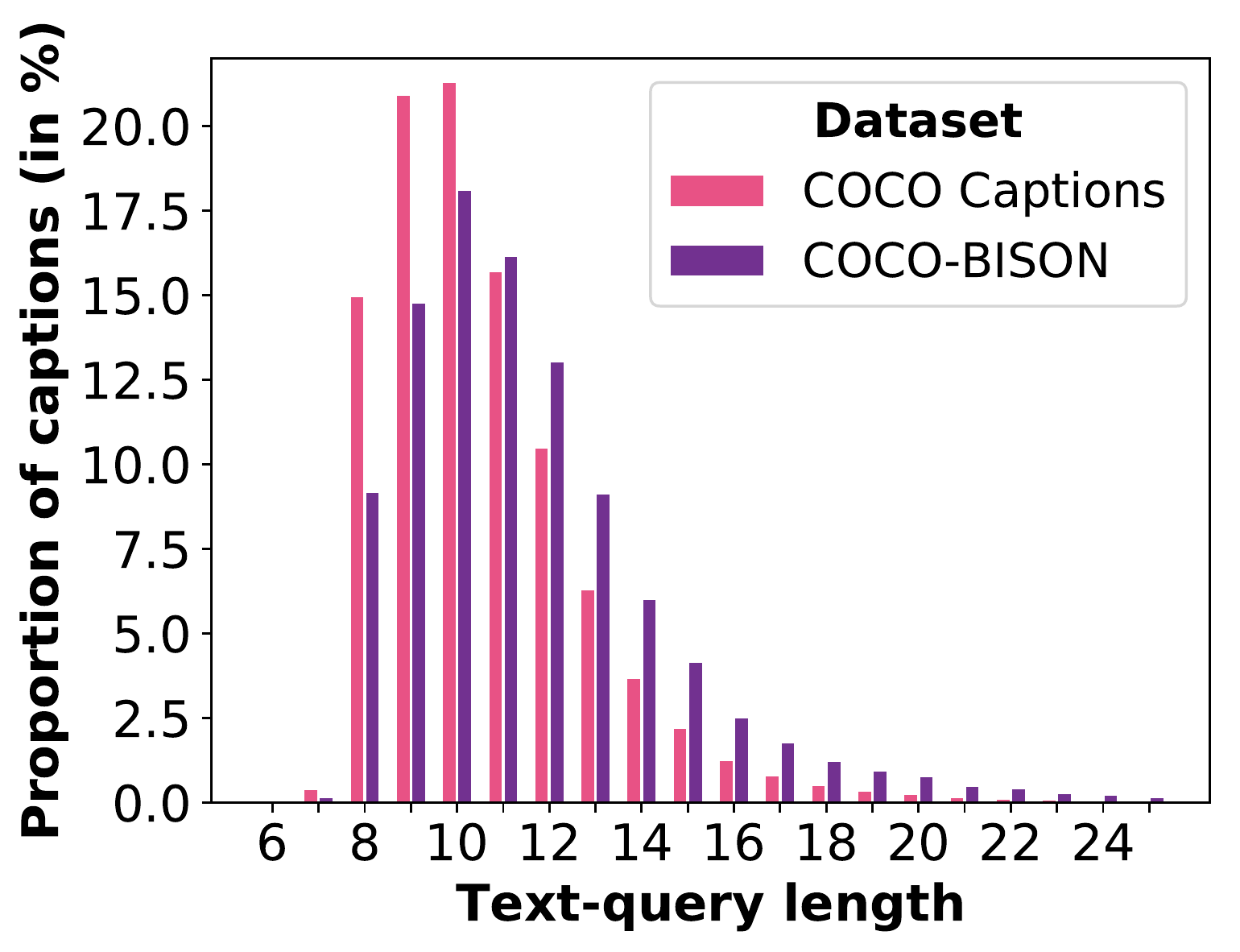}} 
	\subfloat{\includegraphics[width=0.49\linewidth]{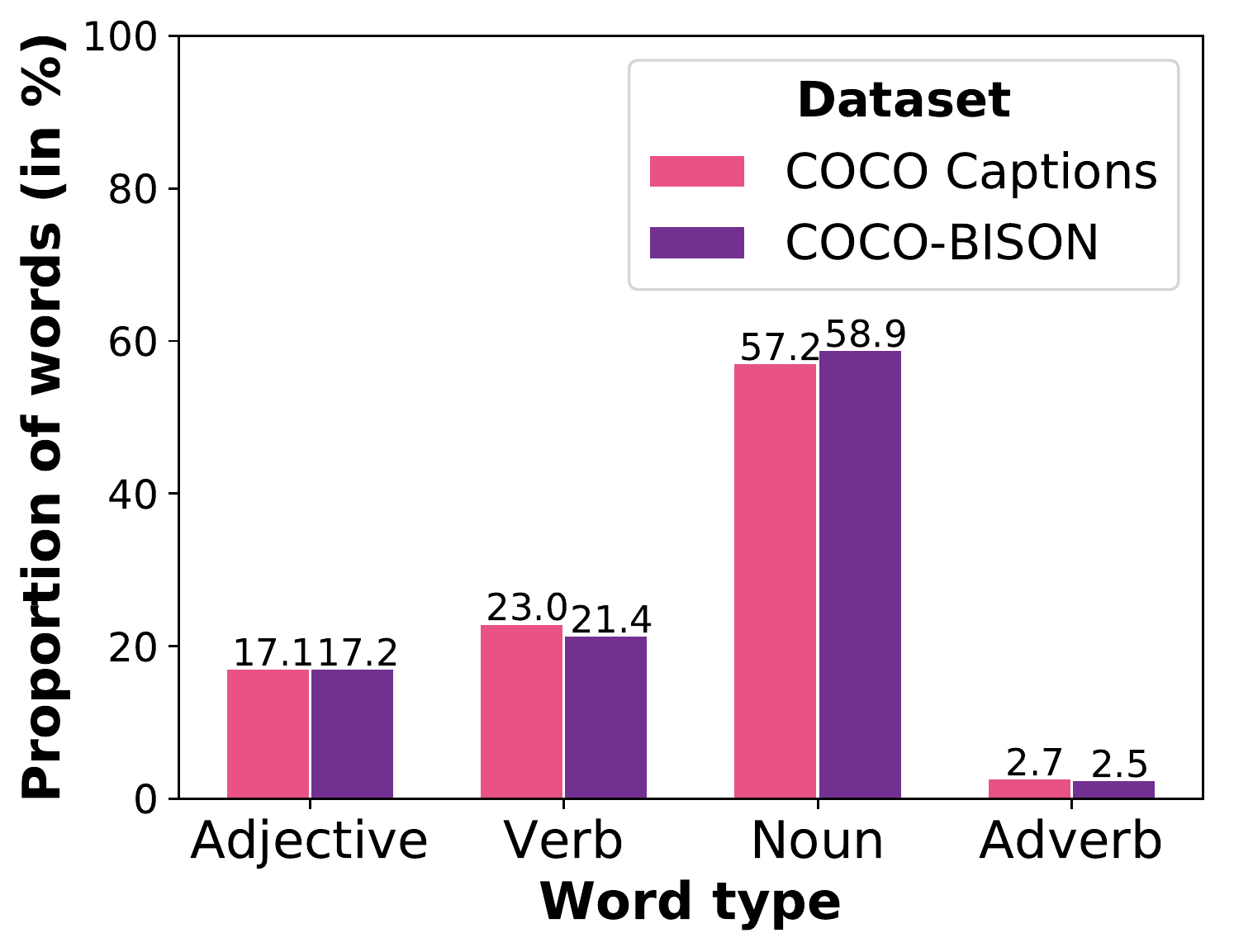}}
	\caption{\textbf{Statistics of COCO Captions and COCO-BISON:} Text-query length distribution (left) and part-of-speech distribution (right) of captions in the datasets. }
	\label{fig:bison_characteristics}
\end{figure}

To ensure our collection procedure did not substantially alter the text distribution of the COCO dataset, we compare\footnote{The word distribution and vocabulary overlap of both datasets is shown in the supplementary material.} the COCO and the COCO-BISON datasets in terms of caption / text query length and part-of-speech distribution in Figure~\ref{fig:bison_characteristics}. The figure shows that, on average, COCO-BISON queries tend to be slightly longer than COCO captions: annotators selected the longest captions $\sim\!30\%$ of the time in the second stage of the dataset collection, presumably, because longer queries tend to be more detailed. However, the part-of-speech distribution of the COCO-BISON queries is very similar to that of COCO captions, facilitating experiments in which image retrieval and captioning systems are trained on COCO but evaluated on COCO-BISON.

\subsection{Definition of the BISON Task}
In the BISON task, the model is given two images and a text query that describes one of the two images and asked to select the correct image; see Figure~\ref{fig:teaser}. The model's performance is measured in terms of binary classification accuracy. We report the mean accuracy over the \OurDataset data and refer to it as the BISON score. We only use \OurDataset for evaluation, \ie, we do not train systems on the annotations in the \OurDataset dataset.

Existing text-based image retrieval and image captioning systems can be used to perform binary image selection. Doing so requires computing a ``compatibility'' score between the text query and the two images, and picking the image with the highest score. For image captioning systems, the compatibility score is generally defined as the log-likelihood of the text query given the image. Image retrieval systems naturally compute the compatibility score, \eg, via an inner product of the image and text features.

%% file: system_benchmarking.tex

\section{BISON Evaluation of State-of-the-Art\\Captioning and Retrieval Systems}
\label{sec:exp_setup}
We evaluate four state-of-the-art text-based image retrieval systems and three recent image captioning systems on binary image selection using the COCO-BISON dataset.

\subsection{Evaluated Retrieval and Captioning Systems}

We evaluate four systems for \textbf{text-based image retrieval}: (1) ConvNet+BoW,
(2) ConvNet+Bi-GRU, (3) Obj+Bi-GRU, and (4) SCAN~\cite{lee2018stacked}.
The \emph{ConvNet+BoW} system represents the text query by averaging word embeddings over all words in the query, and represents the image by averaging features produced by a convolutional network over regions (described later). The resulting representations are processed separately by two multilayer perceptrons (MLPs). We use the cosine similarity between the outputs of the two MLPs as the image-text compatibility score. 
The \emph{ConvNet+Bi-GRU} system is identical to the previous system, but it follows~\cite{Kiros2014UnifyingVE} and uses a bi-directional GRU~\cite{chung2014empirical} to construct the text representation. 
The \emph{Obj+Bi-GRU} system is similar to ConvNet+Bi-GRU but uses a Bi-GRU to aggregate image-region features (spatial ConvNet features or object proposal features) and construct the image representation. Finally, \emph{SCAN}~\cite{lee2018stacked} is a state-of-the-art image-text matching system based on image-region features and stacked cross-attention; we implement two variants of this system, \emph{viz.} one that uses image-to-text (i2t) attention and one that uses text-to-image (t2i) attention. All retrieval systems are trained to minimize a max-margin loss~\cite{faghri2018vse++}. 

We also evaluate three \textbf{image captioning systems}: (1) the \emph{ShowTell} captioning system~\cite{vinyals2015show}; (2) an extension of the ShowTell system that can attend to specific parts of the image, called \emph{ShowAttTell}~\cite{rennie2017self,Xu2015ShowAA}; and (3) the state-of-the-art \emph{UpDown} captioning system~\cite{Anderson2017BottomUpAT}. Like ShowAttTell, the UpDown system uses a spatial attention mechanism but it differs from ShowAttTell in that it uses two LSTMs: one for decoding captions and another one for generating spatial attention over image features. We train all three captioning systems on the COCO captions~\cite{chen2015microsoft} training set by minimizing the cross-entropy loss per word over a vocabulary of $9,487$ words, and average the loss over all words in the reference caption. Following common practice~\cite{Anderson2017BottomUpAT,luo2018discriminability,rennie2017self}, we also finetune the systems using self-critical sequence training (SCST;~\cite{rennie2017self}). SCST uses the REINFORCE algorithm~\cite{sutton2000policy} to maximize the CIDEr-D score~\cite{vedantam2015cider} of the captioning system. We report the performance of the captioning systems both before and after SCST finetuning.

\par \noindent \textbf{Implementation details.} Following the current state-of-the-art in image captioning~\cite{Anderson2017BottomUpAT,lee2018stacked}, all our systems use the top $36$ object proposal features produced by a Faster R-CNN model~\cite{ren2015faster} with a ResNet-101 backbone that was trained on the ImageNet~\cite{russakovsky2015imagenet} and Visual Genome~\cite{krishna2017visual} datasets. In all systems, word embeddings were initialized randomly. We refer the reader to the supplementary material for a complete overview of the hyperparameters we used when training the systems. 

\begin{table}[t]
	\tabcolsep 3pt
	\centering
	\resizebox{\linewidth}{!}{ 
	\begin{tabular}{lccccc}\toprule
		Dataset $\rightarrow$ & \multicolumn{4}{c}{COCO-1K~\cite{karpathy2015deep}} & \OurDataset \\
		Task $\rightarrow$ & \multicolumn{2}{c}{\bf Image retrieval} & \multicolumn{2}{c}{\bf Caption retrieval} & \\
		Measure $\rightarrow$ &  \bf R@1 & \bf R@5 & \bf R@1 & \bf R@5 & \bf \GCSI \\ \midrule
		ConvNet+BoW & 45.19 & 79.26 & 56.60 & 85.70 & 80.48 \\
		ConvNet+Bi-GRU~\cite{Kiros2014UnifyingVE}  & 49.34 & 82.22 & 61.16 & 89.02 & 81.75 \\
		Obj+Bi-GRU  & 53.97 & 85.26 & 66.86 & 91.40 & 83.90 \\
		SCAN i2t~\cite{lee2018stacked} & 52.35 & 84.44 & 67.00 & 92.62 & 84.94 \\
		SCAN t2i~\cite{lee2018stacked} & \bf 54.10 & \bf 85.58 & \bf 67.50 & \bf 92.98 & \bf 85.89 \\ \midrule
	\end{tabular}
	}
	\caption{\textbf{Performance of text-based image retrieval systems:} Recall@$k$ (with $k\!=\!1$ and $k\!=\!5$) of caption-based image retrieval and image-based caption retrieval on the COCO-1K dataset (left) compared to the BISON accuracy on the COCO-BISON dataset (right). See text for details.
	}\label{tab:retrieval_eval}
\end{table}

\subsection{Results}
Table~\ref{tab:retrieval_eval} presents the BISON accuracy of the \textbf{text-based image retrieval systems} on the COCO-BISON dataset. For reference, the table also presents the recall@$k$ (for $k\!=\!1$ and $k\!=\!5$) of these systems on a caption-based image retrieval and an image-based caption retrieval task; these results were obtained on the COCO-1K split of~\cite{karpathy2015deep}. In line with prior work~\cite{lee2018stacked}, we find that the SCANt2i system outperforms the competing systems in terms of all quality measures. 

As expected, we observe that the ranking of caption-based retrieval systems in terms of their BISON accuracy is similar to their ranking in terms of retrieval measures. However, the BISON score provides a more reliable error measure because it does not erroneously consider correct retrievals to be incorrect just because another image happened to be labeled as the positive image for that query. This is reflected in the fact that the BISON score of all systems is higher than their recall@1, and implies that BISON scores are more reliable. We expect that the reliability of evaluation measures becomes more important as the quality of text-to-image matching systems increases: as error rates decrease, measures with both low variance and bias become essential to reliably compare systems.

\begin{figure}[t]
	\centering
	\includegraphics[width=\linewidth]{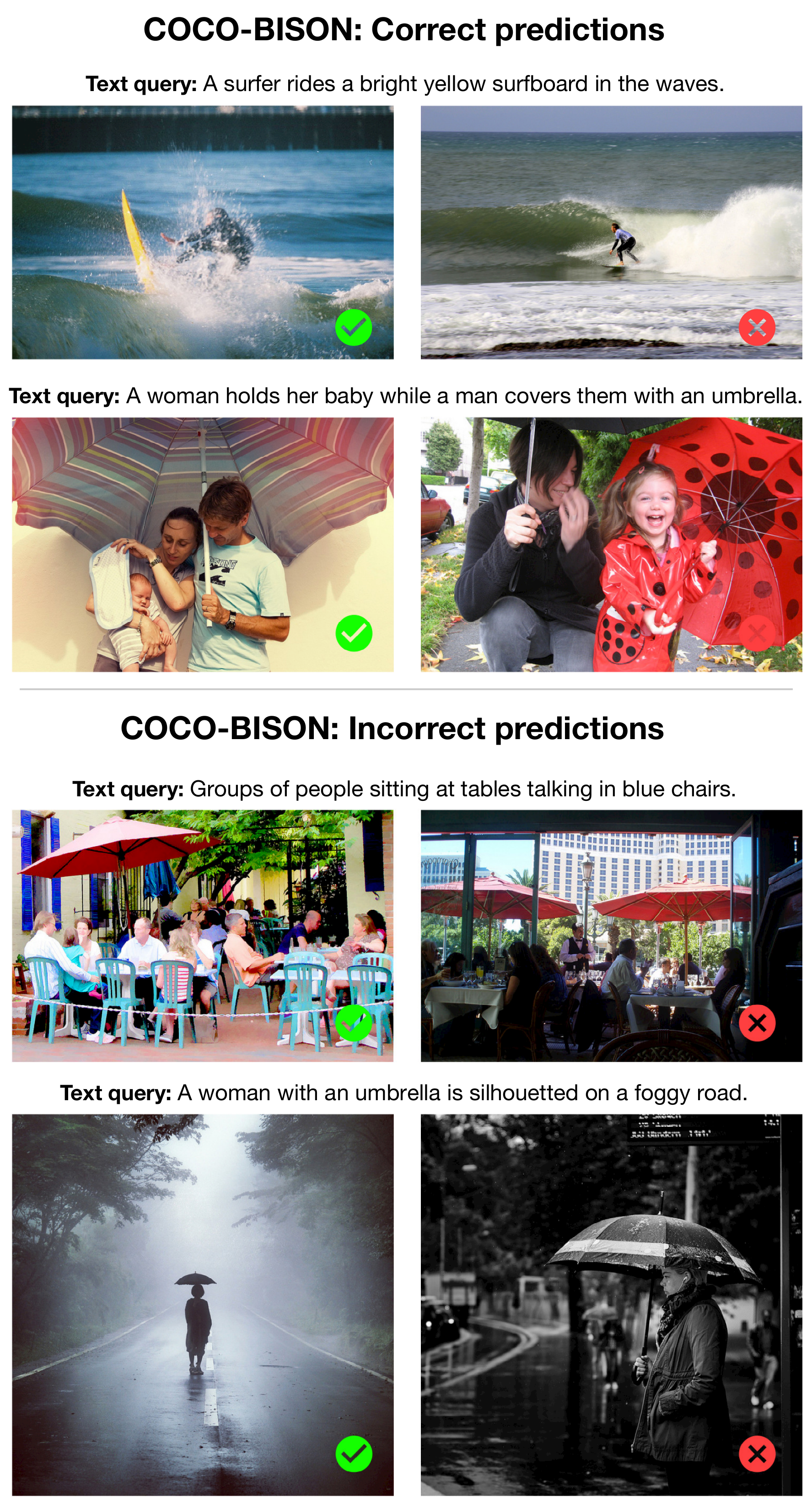}
	\caption{\textbf{Examples of COCO-BISON queries for which correct predictions} (top) \textbf{and incorrect predictions} (bottom) \textbf{were made:} Predictions were performed by scoring both images using the SCAN t2i~\cite{lee2018stacked} retrieval system and selecting the highest-scoring image. The examples illustrate the strong performance of the system, but also show how it may fail to consider fine-grained visual content.}
	\label{fig:bison_example_classifications}
\end{figure}

Figure~\ref{fig:bison_example_classifications} displays two correct and two incorrect predictions for BISON examples, obtained using the SCAN t2i retrieval system. The examples illustrate the strong performance of the system, but also highlight how it sometimes fails to incorporate fine-grained visual content in its predictions, such as the color of the chairs or the presence of fog.

Table~\ref{tab:captioning_eval} presents the BISON accuracy of our three \textbf{image captioning} systems on the COCO-BISON dataset. For reference, the table also presents the performance of these systems in terms of four standard captioning scores on the standard COCO validation set, and the performance of human annotators on the COCO test set (adopted from~\cite{lsun-coco-workshop}). Again, the results reveal that the ranking of the three systems is identical across all evaluation scores, even though BISON measures different aspects of the system than the captioning scores. In line with prior work~\cite{shuster2018engaging}, we find that the UpDown captioning system outperforms its competitors in terms of all evaluation measures, including BISON.

The main difference between BISON and existing captioning scores is in how they rank the ability of humans to generate captions: all three systems outperform humans in terms of nearly all captioning scores, but they all perform substantially worse than humans in terms of BISON accuracy\footnote{Please note that the accuracy of humans on the BISON task is $100\%$ by definition due to the way the COCO-BISON dataset was collected.}. Unless one believes that current image captioning systems actually exhibit super-human performance, this suggests that measuring the BISON score of a system provides a more realistic assessment of the capabilities of modern image captioning systems compared to humans.

\begin{table}[t]
	\tabcolsep 3pt
	\centering
	\resizebox{\linewidth}{!}{  
		\begin{tabular}{lccccc}\toprule
			Dataset $\rightarrow$& \multicolumn{4}{c}{COCO validation split} &  \OurDataset  \\
			Measure $\rightarrow$& \bf BLEU-4 & \bf CIDEr & \bf SPICE & \bf METEOR & \bf \GCSI  \\ \midrule
			\multicolumn{6}{l}{\bf Cross-entropy loss}  \\ \midrule
			ShowTell~\cite{vinyals2015show} & 32.35	& 97.20 & 18.34 & 25.51  & 78.59 \\
			ShowAttTell~\cite{Xu2015ShowAA} & 33.49	& 101.55 & 19.16 & 26.06 & 82.04 \\
			UpDown~\cite{Anderson2017BottomUpAT}  & 34.53 & 105.40 & 19.86 & 26.69 & 84.04 \\ \midrule
			
			\multicolumn{6}{l}{\bf Self-critical sequence loss~\cite{rennie2017self}}  \\ \midrule
			ShowTell~\cite{vinyals2015show}  & 32.38	& 97.88 & 18.42	& 25.68 & 78.79 \\
			ShowAttTell~\cite{Xu2015ShowAA}  & 33.99 & 103.68 & 19.53 & 26.37 & 82.73 \\
			UpDown~\cite{Anderson2017BottomUpAT}  & \textbf{34.58} & \textbf{106.30} & \textbf{20.01} & \textbf{26.92} & 84.27 \\ \midrule			
			 
			Human~\cite{lsun-coco-workshop} & $21.7^{\star}$ & $85.4^{\star}$ & $19.8^{\star}$ & $25.2^{\star}$ & \bf 100.00 \\ \bottomrule
		\end{tabular}
	}
	\caption{\textbf{Performance of image captioning systems:} Four captioning scores measured on the COCO validation set (left) compared to the BISON accuracy on the COCO-BISON dataset (right). Human performances marked with $^\star$ were measured on the COCO test set. See text for details.}\label{tab:captioning_eval}
\end{table}

%% file: related.tex
\section{Related Work}
\label{sec:related_work}

BISON is related to a variety of different tasks and experimental setups that involve matching visual and linguistic information, including referring expressions~\cite{kazemzadeh2014referitgame,kong2014you,yu2016modeling}, visual story-telling~\cite{huang2016visual}, visual question answering~\cite{antol2015vqa,malinowski2015ask}, visual question generation~\cite{du2017learning,misra2017learning,mostafazadeh2016generating}, and zero-shot learning~\cite{akata2013label,romera2015embarrassingly}. A comprehensive overview of all this prior work is outside the scope of this paper; we refer the reader to~\cite{ferraro2015survey} for a survey. Most of the related tasks are more ``AI-complete'' than binary image selection in the sense that they simultaneously assess a range of system abilities that go beyond matching visual content and textual descriptions. 

BISON is most closely related to text-based image retrieval, image captioning, and referring expression tasks. We give a brief overview of prior work on these tasks below.

\par \noindent \textbf{Text-based image retrieval}~\cite{barnard2003matching,barnard2001learning,gong2014multi,hodosh2013framing,plummer2015flickr30k,socher2014grounded} is a task in which the system is asked to retrieve relevant images given a text description. Retrieval performance is generally measured in terms of recall@$k$~\cite{hodosh2013framing}.  Similar to BISON, caption-based image retrieval evaluates how well a system can distinguish relevant images from irrelevant ones. As described above, the key difference between image retrieval and BISON is that retrieval evaluations rely on ``implicit'' negatives: retrieval datasets provide manually annotated positive image-description pairs, but they assume that every image-description pair that is not in the dataset is a negative example. In practice, this assumption is often violated: many such image-description pairs would actually be labeled positively by a human annotator~\cite{li2017ngrams}. In contrast to retrieval datasets, each example in our COCO-BISON dataset contains a positive and a \emph{genuinely} negative image-description pair, which facilitates more reliable evaluation.

\par \noindent \textbf{Image captioning}~\cite{aneja2018convolutional,chen2015mind,fang2015captions,hodosh2013framing,karpathy2015deep,kulkarni2013babytalk,vinyals2015show,Xu2015ShowAA} is a task in which the system generates a textual description of an image. The task assesses a system's ability to ingest visual information in an image and generate fluent natural language descriptions of that information~\cite{chen2015microsoft}. As a result, captioning gauges not only visual understanding but also the generative linguistic prowess of systems. In contrast, binary image selection only measures the ability of a system to discriminate between images based on a text description. 

A recent line of work~\cite{andreas2016reasoning,luo2018discriminability,vedantam2017context} focuses on generating discriminative text descriptions for images. BISON is related to this work but it focuses solely on the discriminative aspect of the task by not considering language generation. This relates BISON to~\cite{suhr2018corpus}, which centers on predicting whether or not text correctly describes an image pair.


\par \noindent \textbf{Referring expressions}~\cite{dale1995computational,kazemzadeh2014referitgame,kong2014you,mitchell2012midge,yu2016modeling} is a task in which a system is asked to distinguish objects in a \emph{single} image based on a text description. BISON can thus be viewed as a kind of \emph{holistic} referring-expressions task that involves between-image rather than within-image comparisons. In contrast to referring expressions that focus on a single object and its attributes, text descriptions in BISON may focus on groups of objects and their attributes, relationships between these objects, and even entire scenes. 

%% file: bison_analysis.tex
\section{Using BISON to Analyze Systems}
\label{sec:bison_analysis}

We performed experiments to study BISON evaluation and the differences between text-based image retrieval and image-captioning systems in more detail.

\begin{figure}[t]
	\centering
		\subfloat{\includegraphics[width=0.512\linewidth]{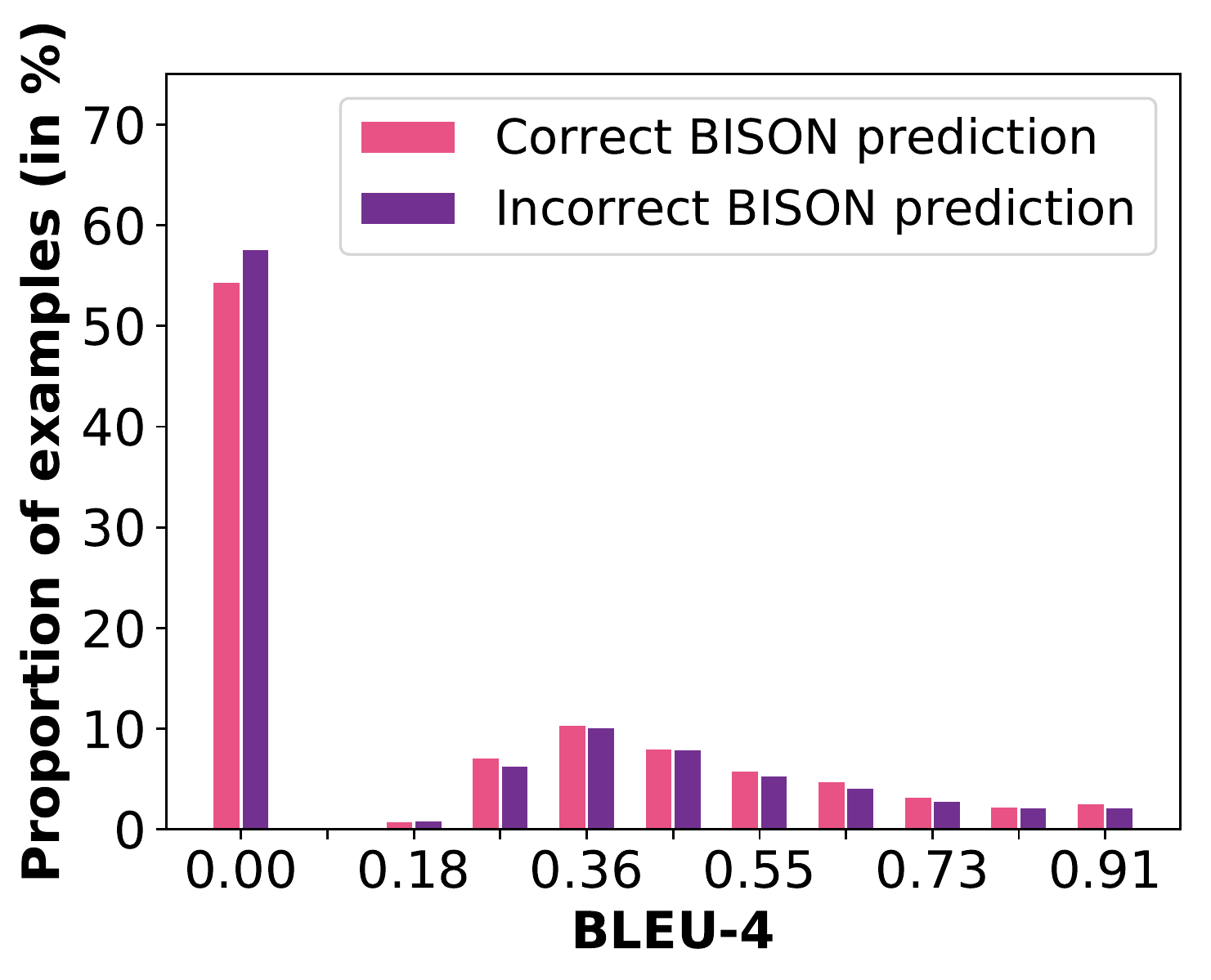}} 
		\subfloat{\includegraphics[width=0.488\linewidth]{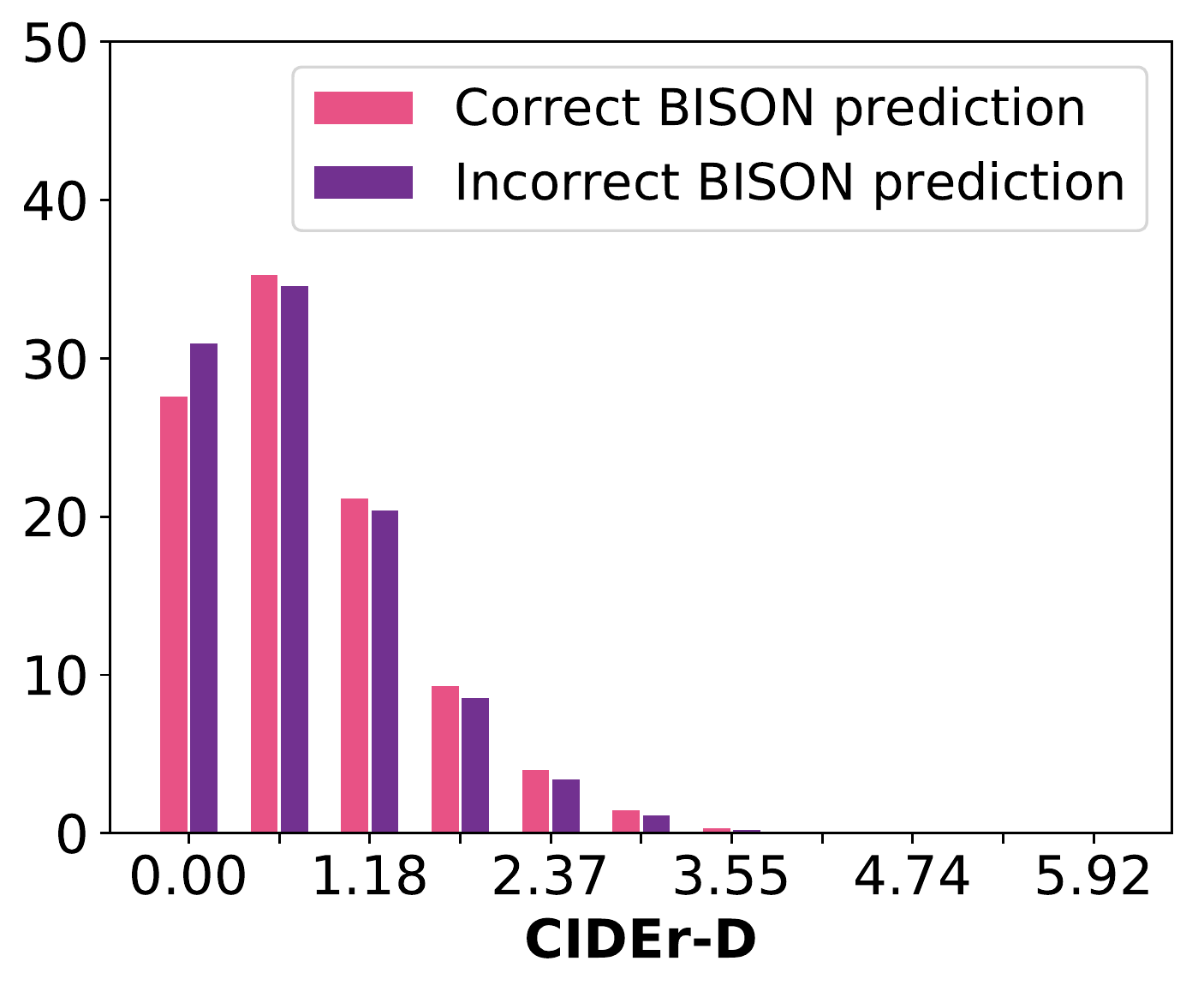}} \\
		\subfloat{\includegraphics[width=0.512\linewidth]{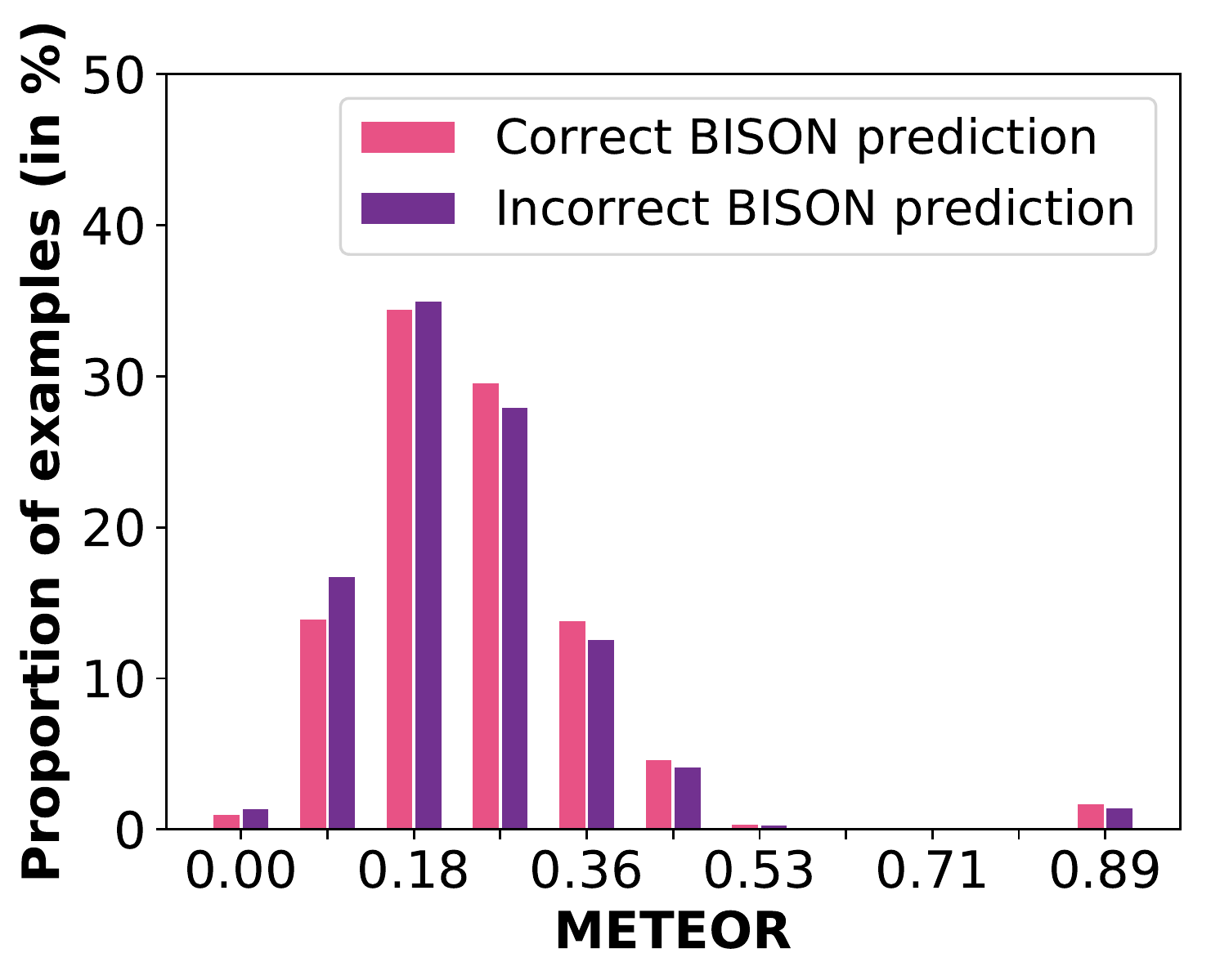}}
		\subfloat{\includegraphics[width=0.488\linewidth]{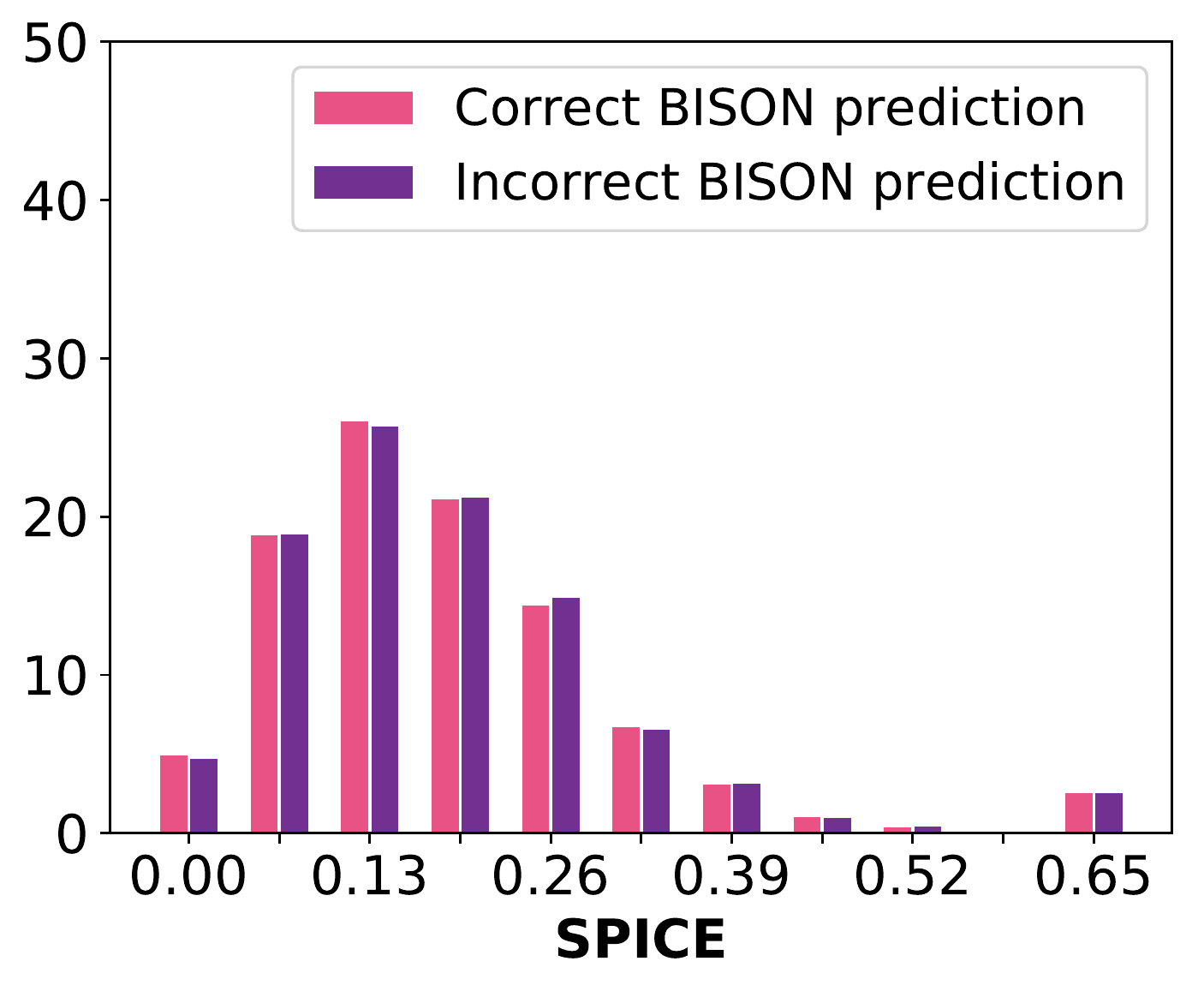}}
	\vspace{-0.05in}
	\caption{\textbf{Distribution of captioning scores} (BLEU-4, CIDEr, METEOR, and SPICE) \textbf{for correct and incorrect BISON predictions:} BISON predictions were made using the UpDown captioning system~\cite{Anderson2017BottomUpAT}, and captioning scores were measured using the same system. The captioning-score distribution for BISON examples that were correctly predicted is shown in pink; the distribution for incorrect BISON predictions is shown in purple.  The captioning-score distribution is nearly identical between correct and incorrect BISON predictions, suggesting that these scores measure something very different than BISON accuracy.}
	\label{fig:hist_caption_score_vs_bison}
\end{figure}

\par \noindent \textbf{Does BISON accuracy predict captioning scores?}
We evaluated the UpDown captioning system~\cite{Anderson2017BottomUpAT} in terms of BISON accuracy and in terms of four captioning scores on the COCO-BISON dataset. Figure~\ref{fig:hist_caption_score_vs_bison} shows the distribution of the captioning scores \emph{separately} for BISON examples that were classified correctly and for examples that were incorrectly (by the same systems). Specifically, for each \OurDataset example that the model classifies correctly (or incorrectly), we generated a caption for the \emph{positive} image and measured the captioning score of the generated caption. If captioning scores measure the same characteristics as BISON, one would expect the distribution of captioning scores to center on higher values for correctly classified BISON examples than for incorrectly classified BISON examples. However, the figure shows that distribution of all the captioning scores is nearly identical for BISON examples that were correctly and incorrectly classified. This result suggests that BISON assesses different aspects of matching visual and linguistic content than image captioning does.

\par \noindent \textbf{Do caption score differences provide signal for BISON?}
We try and classify BISON examples based on captioning scores for the \emph{positive} and \emph{negative} images in those examples. Specifically, we compute a captioning score (\eg, BLEU-4) between the BISON text query and the COCO reference captions corresponding to both the \emph{positive} and \emph{negative} BISON images. When computing the score for the positive image, we remove the text query from the set of reference captions; for the negative image, we randomly select four captions from the reference captions (without replacement). Next, we select the image with the highest captioning score as prediction for the BISON example.

The BISON accuracy of this approach is $70.73\%$ for BLEU-4, $70.78\%$ for CIDEr-D, $74.44\%$ for METEOR, and $62.79\%$ for SPICE. This result shows that predictions based on captioning scores select the incorrect BISON image at least $25\%$ of the time, despite relying on access to the ground-truth reference captions. This low accuracy suggests that BISON and captioning scores are very different measures for matching textual and visual content.

\begin{figure}[t]
	\centering
	\includegraphics[width=0.8\linewidth]{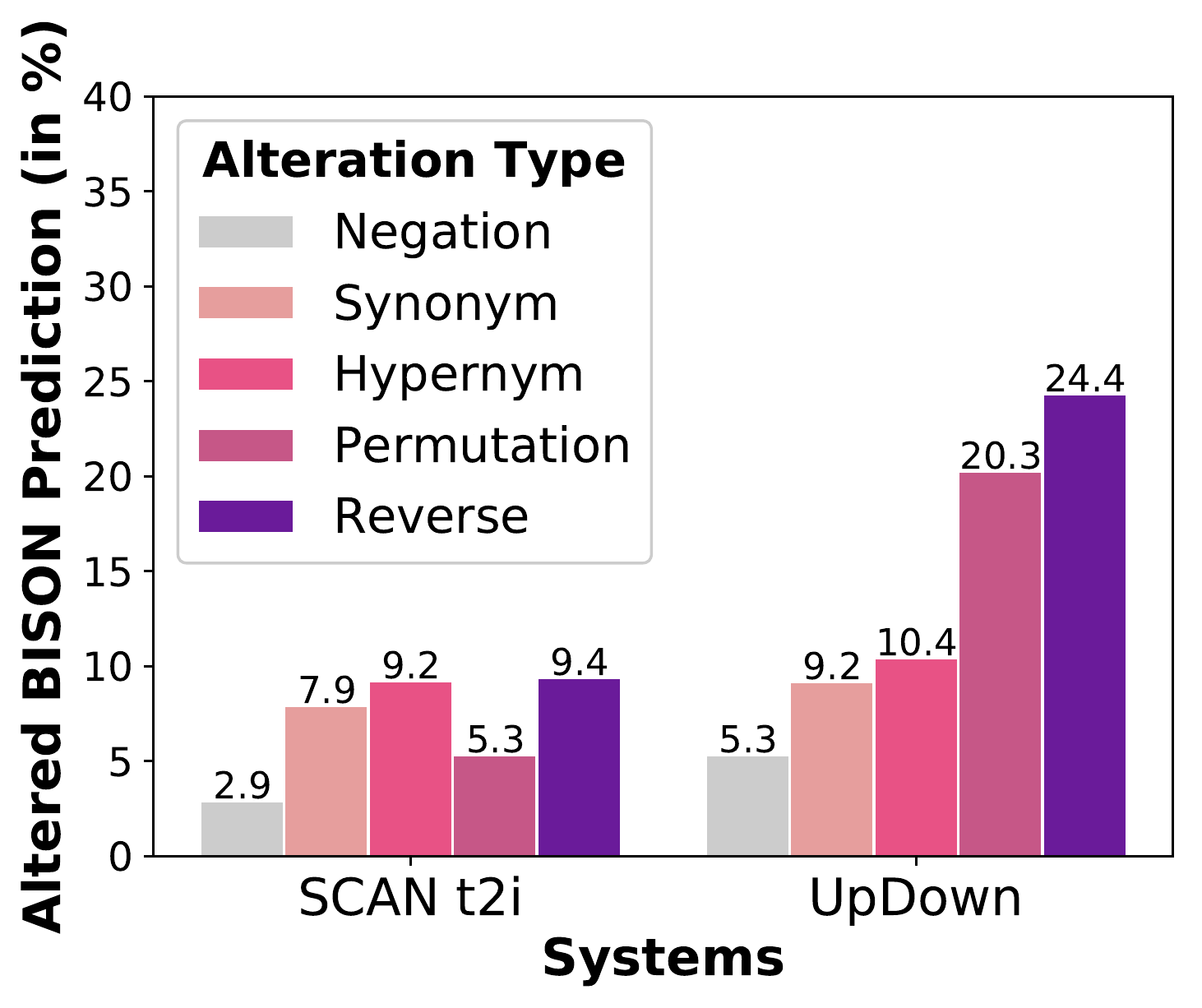}
	\caption{\textbf{Changes in BISON prediction under automatic text-query perturbations.} 
	Percentage of predictions by the SCAN t2i retrieval and UpDown captioning systems on COCO-BISON that are changed when words in the query are permuted or the query is negated \etc. Results show that the retrieval model largely ignores word order, and both models largely ignore negations.}
	\vspace{-0.05in}
	\label{fig:alter_text}
\end{figure}

\par \noindent \textbf{How do text-query alterations alter BISON predictions?} We performed experiments where we altered the COCO-BISON text queries by doing the following automatic perturbations: (1) negating the query; replacing the words by (2) synonyms (3) hypernyms; (4) permuting the word order and (5) reversing the word order in the query (see details in the supplementary material). We measured the percentage of BISON predictions that were changed by these query perturbations for both the retrieval and captioning systems. These changes give more insights into the differences between retrieval and captioning systems.
Figure~\ref{fig:alter_text} shows that the BISON predictions of retrieval and captioning systems do not change much under negation. Retrieval systems are also more robust to changes in the word order than captioning systems. This suggests that captioning systems are more sensitive to the fluency of a text query. We report this analysis for other systems in the supplementary material.


%% file: conclusion.tex

\section{Discussion}
This study has developed binary image selection (BISON) as an alternative task for evaluating the performance of systems that relate visual and linguistic content. Our study shows that BISON solves the issues of text-based image retrieval tasks that erroneously assume that all unlabeled images are negative examples for the text query, and that it assesses a different set of capabilities than image captioning tasks. Compared to text-based image retrieval, BISON has the advantage that the evaluation is more reliable, easily interpretable, and that it focuses more on ``fine-grained'' visual content. This focus on fine-grained visual information is also in contrast to image captioning tasks that encourage the generation of ``generic'' descriptions. In a sense, BISON can be viewed as a variant of referring-expressions tasks that considers images ``holistically'' rather than focusing on image parts. However, the BISON paradigm also has disadvantages compared to tasks such as image captioning: for instance, it does not assess the fluency of the linguistic content. Therefore, we view binary image selection as an evaluation task that ought to be used \emph{in conjunction} with other evaluation tasks, such as the image retrieval, image captioning, and referring expression tasks (see Section~\ref{sec:related_work} for an overview).

We observed that the relative ranking of modern systems in terms of retrieval or captioning scores is nearly identical to the ranking of those systems in terms of BISON. A potential explanation for this observation may be that some systems are unequivocally better than others: for instance, if system A has an image-recognition component that is substantially better than the image-recognition component of system B, it is quite likely that system A will outperform system B in a very wide range of tasks involving vision and language. We do emphasize, however, that it is well possible that the observed rank correlation between captioning and BISON scores may no longer hold when researchers start designing systems with the BISON evaluation in mind. Results comparing the performance of humans with that of our systems underline this point: existing captioning scores suggest that systems possess super-human capabilities, which contradicts human assessments of the quality of these systems. By contrast, the BISON scores of current systems appear to be better aligned with human assessments of the quality of these systems.

To conclude, we hope that the binary image selection task will foster research into models that go beyond coarse-level matching of visual and linguistic content by rewarding systems that can perform visual grounding at a detailed level. The interpretability of BISON makes it easier to debug and analyze this visual grounding. We hope that the public release of the COCO-BISON dataset will help the community assess whether we are making progress towards the goal of developing such systems.

%% file: supp_content.tex
\section*{Supplementary Material}

\section{Dataset Annotation}
\label{sec:annotation}

We present additional details on the annotation process we used for collecting \OurDataset. We provided an overview of this process in Section 4.1 of the main paper.

\subsection{Annotation Interfaces} 
\begin{figure}[h]
	\centering
	\includegraphics[width=\linewidth]{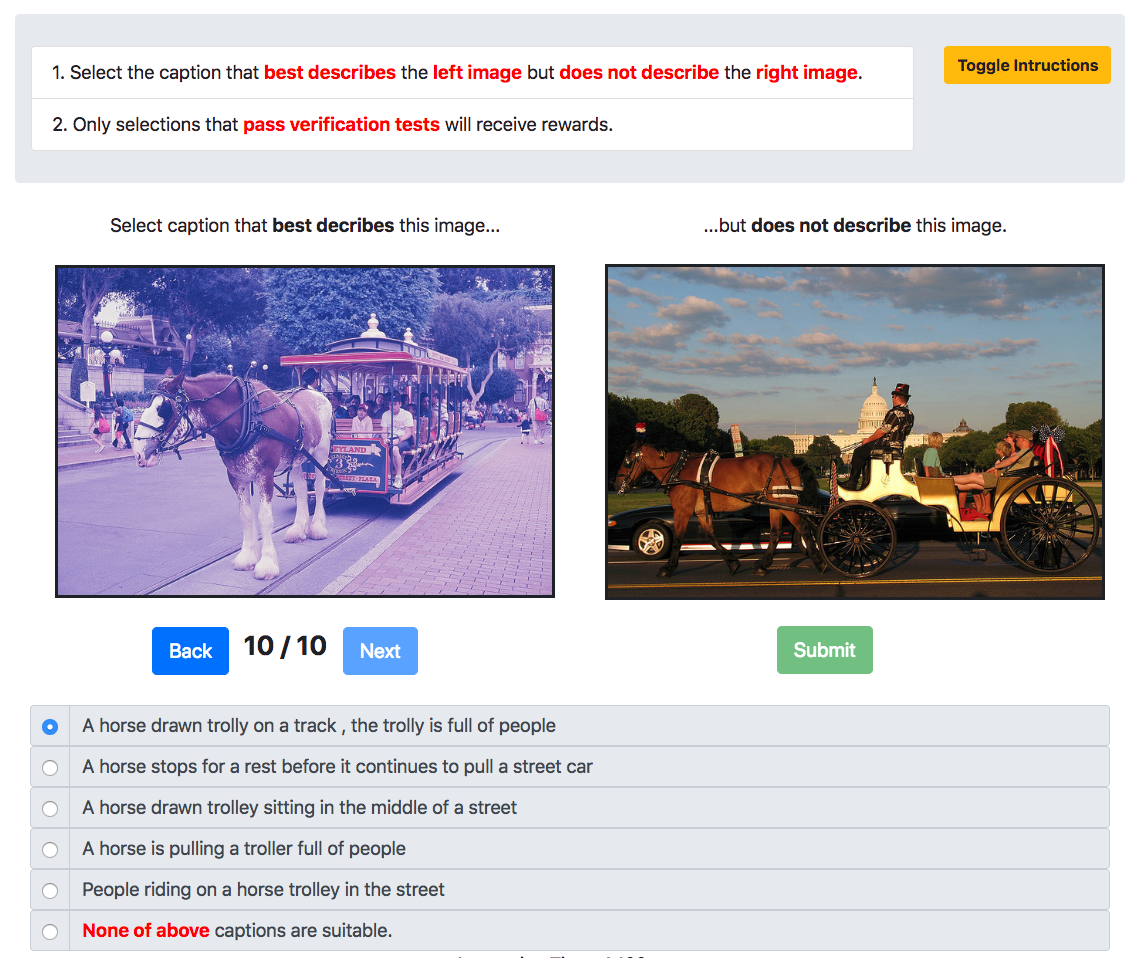}
	\caption{Annotation UI for  \OurDataset: finding \textbf{the most discriminative caption} between a given pair of positive and negative image.}
	\label{fig:ui_annotate}
\end{figure}

\begin{figure}[h]
	\centering
	\includegraphics[width=\linewidth]{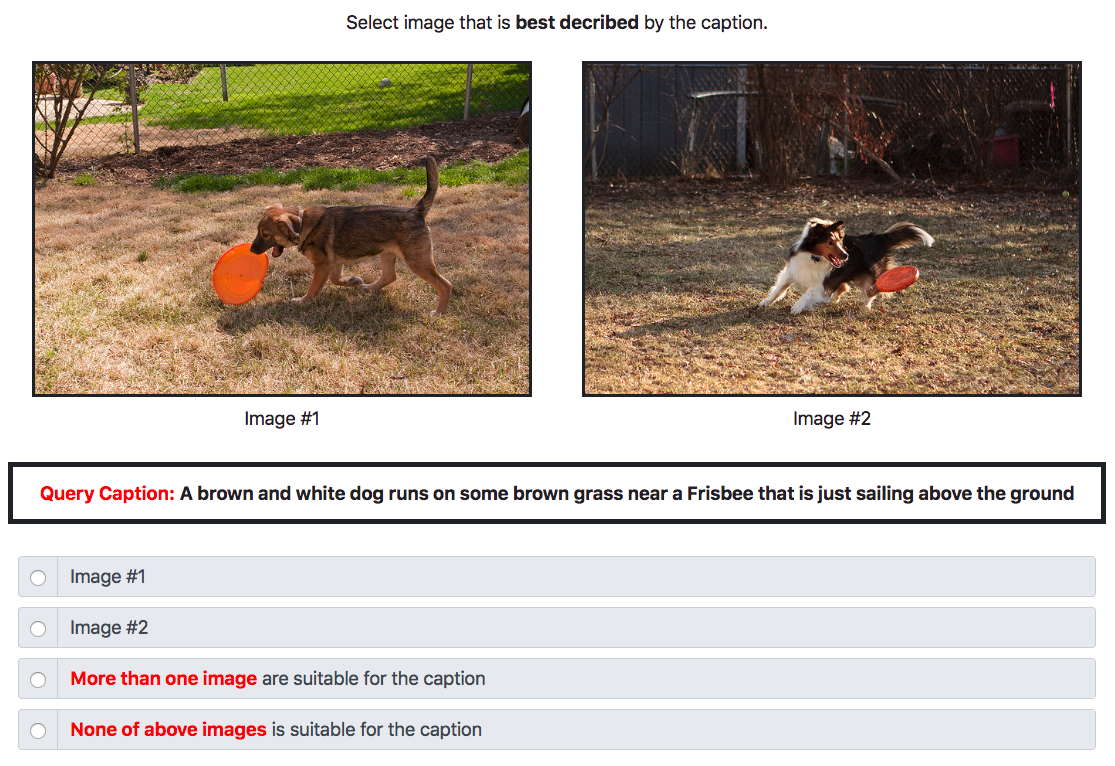}
	\caption{Verification UI for  \OurDataset: selecting the positive image from a pair given the positive caption.}
	\label{fig:ui_verify}
\end{figure}

In \textbf{Stage 1} of the collection of  \OurDataset, we automatically identified pairs of semantically similar images in the COCO validation set (as described in the paper). In \textbf{Stage 2} of the collection process, we asked human annotators to select a caption for each image pair that describes one image (the ``positive'' image) but not the other (the ``negative'' image). Figure~\ref{fig:ui_annotate} shows the annotation interface we used in Stage 2; the annotator used this interface to pick one of the five captions which distinguishes the positive image from the negative image. The five captions shown in the interface are the five captions associated with the positive image in the COCO Captions dataset~\cite{chen2015microsoft}. The interface also provides a ``none of the above'' option that the annotator can select in case none of the five captions is discriminative. In \textbf{Stage 3} of the dataset collection process, we performed verification of the annotations obtain in stage 2, by asking  a new set of annotators to select the image that best matches the caption. The interface for the verification task is shown in Figure ~\ref{fig:ui_verify}; it shows the positive and negative image (in random order) and the caption that the annotator in stage 2 selected for the positive image. The annotations collected in stage 3 are used to verify that the selected text query, indeed, distinguishes between the positive and the negative image.

\subsection{Annotator Qualification Criteria} 
We use publicly available crowdsourcing platform to gather annotations for \OurDataset. To ensure high quality annotations, we defined criteria the annotators must meet in order to contribute to collection of our dataset. Specifically, we require a annotator to have completed at least 500 tasks historically with an acceptance rate over $97\%$, and is from a region that English is the native language. During the annotation process, we group the annotators into two separate groups that were responsible for the second and third stages of the dataset collection, respectively (as described in the paper); annotators could only provide annotations for one of the two stages. This strict separation between annotator groups prevents cases in which an annotator selects a caption in Stage 2 and then verifies their own annotation in Stage 3.

\paragraph{Stage 2 Instructions.} Annotators in stage 2 were presented with the following instructions:

\begin{mdframed}
\begin{enumerate}[leftmargin=*]
	\item \textbf{Submission.} The submission button will be enabled when a caption for all 10 image-pairs is selected.
	\item \textbf{Image dimensions.} Variations in object size due to differences in image dimensions should be considered as irrelevant.
	\item \textbf{Discriminative captions.} (see Figure~\ref{identification_instruction_3} as examples). We perform verification tests on your submission to ensure you selected captions that describe the left (positive) image well, whilst not describing the right (non-positive) image well. Bad caption selections will not be rewarded.
	\item \textbf{What if images are nearly identical, or no suitable caption is available?} Please select the option "None of above captions are suitable". (see Figure~\ref{identification_instruction_4} as example)
	\item \textbf{Images are ``Unavailable''} (see Figure~\ref{identification_instruction_5} as example). Please select the option "None of above captions are suitable".
\end{enumerate}
\end{mdframed}

~\\~\\
\paragraph{Stage 3 Instructions.} Annotators in stage 3 were presented with the following instructions:

\begin{mdframed}
\begin{enumerate}[leftmargin=*]
	\item \textbf{Submission.} The submission button will be enabled when a caption for all 10 image-pairs is selected.
	\item \textbf{Image dimensions.} Variations in object size due to differences in image dimensions should be considered as irrelevant.
	\item \textbf{Most suitable image for a query.} (see Figure~\ref{identification_instruction_3} as examples). We perform verification tests on your submission to ensure you selected captions that describe the left (positive) image well, whilst not describing the right (non-positive) image well. Bad caption selections will not be rewarded.
	\item \textbf{What if there are more than one image that are suitable for the provided text query?} Please select the option "More than one image are suitable for the caption".
	\item \textbf{What if none of images is suitable for the provided text query?} Please select the option "None of above images  is suitable for the caption".
\end{enumerate}
\end{mdframed}

\paragraph{Statistics of Data Collection Process.} After stage 1 of the dataset collection, we obtained 67,564 pairs of a positive and a negative image. In stage 2, the annotators created 61,861 valid query-positive-negative triples, corresponding to a conversion rate of $91.56\%$. For the remaining $8.44\%$ pairs of images, the annotators selected the ``none of the above'' option. In stage 3, we use two separate annotators to verify the query-positive-negative triple and obtained a total of $54,253$ query-positive-negative triples (conversion rate: $87.70\%$) that were confirmed to be correct by \emph{both} the Stage 3 annotators. These triples form the \OurDataset dataset. Table~\ref{tab_worker_stats} summarizes key statistics about the annotators who performed the dataset collection.

\begin{table}[t]
	\centering
	\resizebox{\linewidth}{!}{ 
	\begin{tabular}{lcc}\toprule
		& \bf Stage 2 & \bf Stage 3 \\\midrule
		Number of unique annotators 		       & 173   & 254 \\
		Average time per annotation (in sec.)  & 25.9  & 12.7  \\\bottomrule
	\end{tabular}
	}
	\caption{\textbf{Key statistics of annotators for COCO-BISON dataset} in the stage 2 and stage 3 of dataset collection.}
	\label{tab_worker_stats}
\end{table}  

\section{Details on Analyzing Systems and Measures}

\begin{figure}[h]
	\centering
	\includegraphics[width=\linewidth]{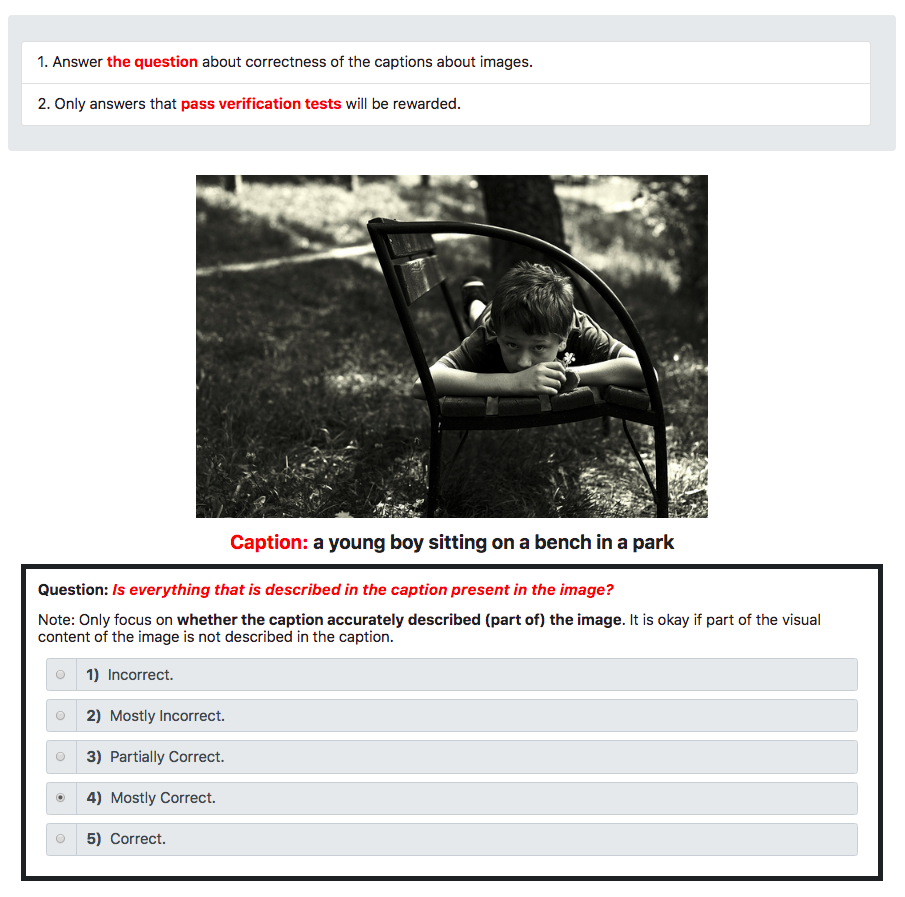}
	\caption{Annotation interface for evaluating the \textbf{correctness} of COCO captions. We followed the COCO guidelines for human evaluation~\cite{lsun-coco-workshop} (human measure 3) to design our user interface.}
	\label{fig:ui_human_correctness}
\end{figure}

\begin{figure}[h]
	\centering
	\includegraphics[width=\linewidth]{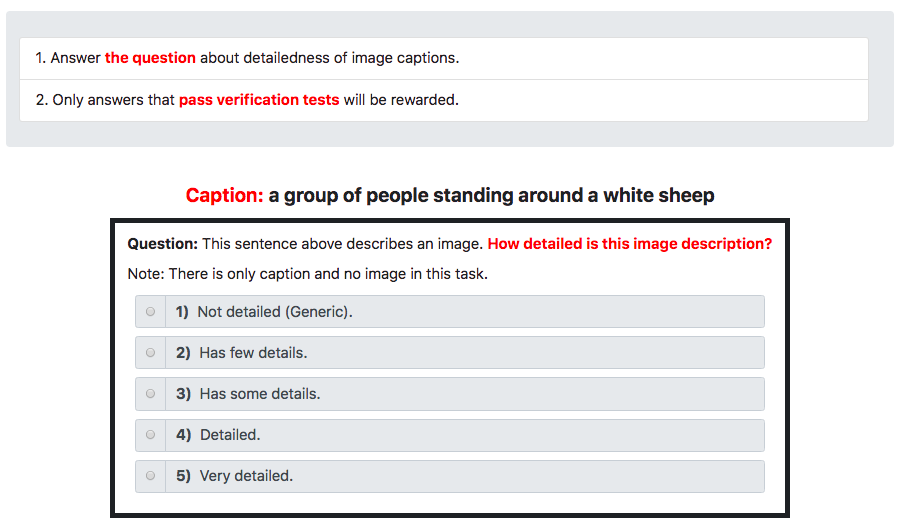}
	\caption{Annotation interface for evaluating the \textbf{detailedness} of COCO captions. We followed the COCO guidelines for human evaluation~\cite{lsun-coco-workshop} (human measure 4) to design our user interface.}
	\label{fig:ui_human_detailedness}
\end{figure}

\paragraph{Annotation Interfaces for Human Evaluation of Generated Captions.} 
As discussed in the main paper (Section 3), we performed human evaluation on the captions generated by the UpDown~\cite{Anderson2017BottomUpAT} system on the COCO validation set. We followed the COCO guidelines for human evaluation~\cite{lsun-coco-workshop} and measure ``correctness'' and ``detailedness''. The corresponding annotation interface we used to gather correctness and detailedness annotations are shown in Figures~\ref{fig:ui_human_correctness} and~\ref{fig:ui_human_detailedness}, respectively. The instructions used in these interfaces were adopted literally from~\cite{lsun-coco-workshop}.

\paragraph{Details on Human Evaluation of Retrieval System.} 

Table 1 in the main paper showed the proportion of ``mistakes'' made by SCAN t2i~\cite{lee2018stacked} retrieval system that human annotator marked as correct. To the details of this study, we asked an expert annotator to example all the ``R@5'' mistakes SCAN t2i system has made (of which the rank of truth example is smaller than 10, to ease of performing annotation). The annotator is presented with all mistakes the system  has made and asked to identify whether there exists a ``mistakes'' is correctly described by the text query.

\section{Comparing the \OurDataset and\\COCO Captions Datasets}
\label{sec:analysis}

\begin{figure}[h]
	\centering
	\includegraphics[width=0.475\textwidth]{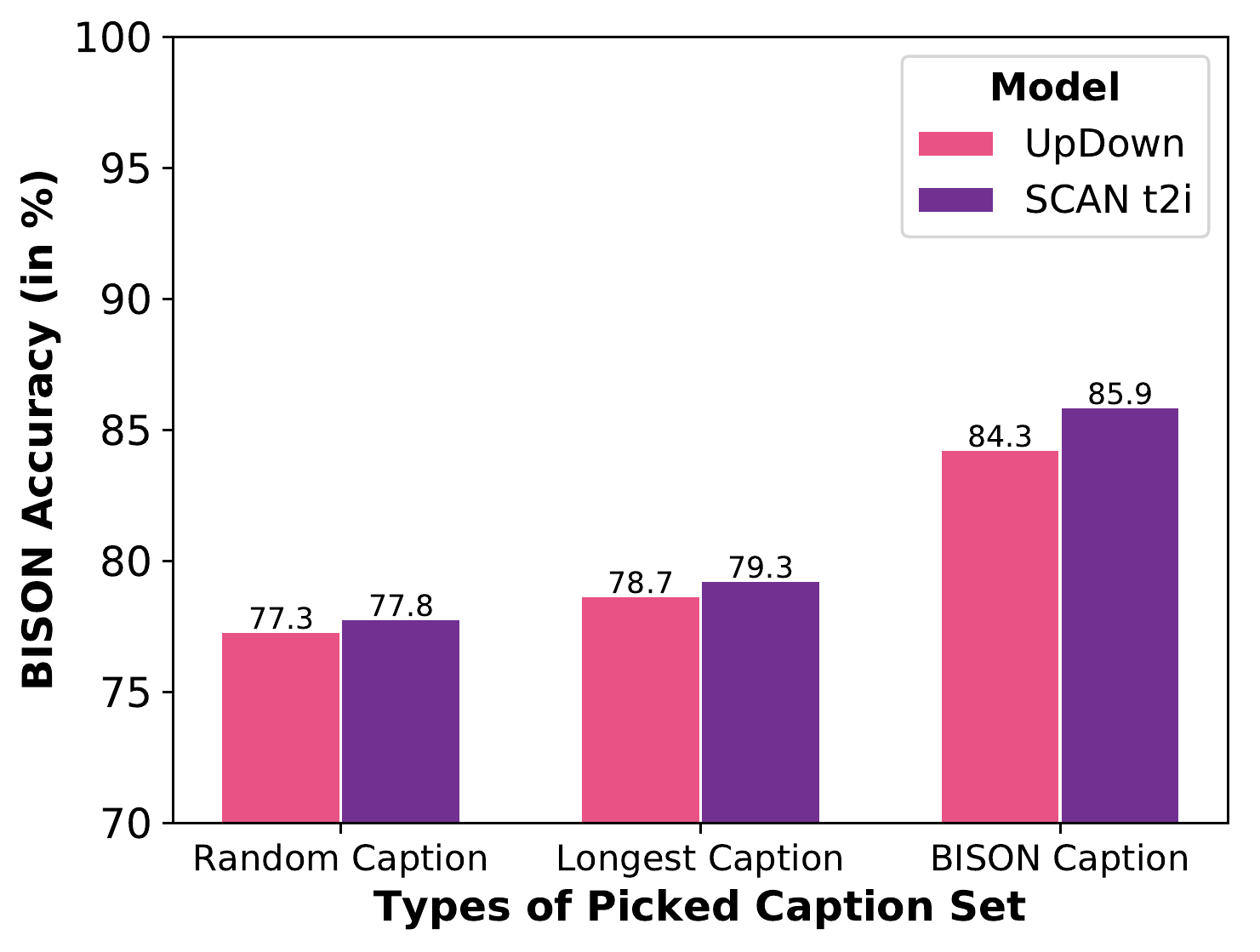}
	\caption{\textbf{BISON performances with differently picked positive caption.} We report the results of representative image captioning system and image retrieval system, UpDown~\cite{Anderson2017BottomUpAT} and SCAN~\cite{lee2018stacked} (t2i).}
	\label{fig_cap_eval}
\end{figure}

\paragraph{Vocabulary Overlap with COCO Captions.} The captions in the  \OurDataset dataset contain a total of 10,800 unique words. By contrast, the COCO Captions validation set contains 18,545 distinct words. Note that, by construction, the vocabulary of \OurDataset is a subset of the vocabulary of COCO Captions. Specifically, \OurDataset contains $58.24\%$ of the COCO Captions vocabulary.

\paragraph{Discriminativeness of \OurDataset Captions.} As described in the main paper, we performed experiments to quantitatively evaluate the ``discriminativeness'' of captions in \OurDataset, compared to the original COCO Captions dataset. Specifically, given all (five) candidate COCO captions associated with an image, we evaluated the discriminativeness of (1) a \emph{Random Caption} and (2) the \emph{Longest Caption} among the five captions. 

Concretely, we train the state-of-the-art caption based image retrieval system SCAN~\cite{lee2018stacked} and image captioning system UpDown~\cite{Anderson2017BottomUpAT} (details in Section~\ref{sec:implementation}) on the standard COCO Captions training set and evaluate model's BISON performance on \OurDataset. The results are shown in Figure~\ref{fig_cap_eval}. The model's matching performance on \OurDataset is $\sim\!7\%$ higher than that of \emph{Random Caption} and \emph{Longest Caption}. This result suggest that given the same model, the captions we  collected in \OurDataset are more discriminating between two semantically similar images. It is also worth noting that, by definition, the human performance on the BISON task is 100\% due to the way the COCO-BISON dataset was collected.

\section{Implementation Details}
\label{sec:implementation}
We now describe the implementation details for the experiments in the main paper.

\subsection{Image Features}
As described in the paper, we follow~\cite{Anderson2017BottomUpAT} to obtain visual features using the object proposals from a Faster-RCNN~\cite{ren2015faster} detection model. The ConvNet backbone is a ResNet-101~\cite{he2016deep} pretrained on ImageNet. We train the detection model on the Visual Genome detection dataset (which contains annotations of $1600$ different objects and $400$ attributes). This detection model is then used to obtain visual features for the input image. Specifically, we resize the short side of a image to $600$ pixels (keeping aspect ratio fixed) and get the $36$ object proposals for which the detector confidence is the highest. We then apply RoI pooling~\cite{ren2015faster} to obtain the averaged $2048$-dimensional object features for each object proposal. This process results in a $36 \times 2048$ dimensional visual feature for each image.

\subsection{Image Captioning Systems}

\paragraph{ShowTell~\cite{vinyals2015show}:} We follow~\cite{vinyals2015show} and implement a LSTM~\cite{hochreiter1997long}-based caption generator, using the following holistic visual feature. We extract am $2048$-dimensional image feature by averaging the $36$ object proposal features that are described above. A linear layer is applied to the resulting feature vector to obtain a $512$-dimensional holistic image feature, which is then fed into the caption-decoding LSTM as its first input word embedding. The parameters in the word embeddings of the LSTM caption decoder are initialized by random sampling from a uniform distribution over the domain $[-0.1, 0.1]$.

\paragraph{ShowAttTell~\cite{Xu2015ShowAA,rennie2017self}:} We follow~\cite{Xu2015ShowAA} and implement a LSTM-based caption generator using the $36 \times 2048$ object proposal features and attention mechanism. The overall architecture is similar to~\cite{Xu2015ShowAA}; we adopt the modification suggested by~\cite{rennie2017self} and input the attention-derived image features to the cell node of the LSTM. Concretely, during the decoding of each words, an attention vector is computed based on the hidden state of the caption-decoding LSTM. The attention weights are then used to compute a weighted combination of the $36 \times 2048$ object proposal features into a $2048$-dimensional feature vector. This averaged feature is then input into the memory cell of the LSTM, in conjunction with the previous word embeddings, to decode the caption~\cite{rennie2017self}.

\paragraph{UpDown~\cite{Anderson2017BottomUpAT}:} We implement the model proposed in ~\cite{Anderson2017BottomUpAT} using two LSTMs: one for generating top-down attention that combines object proposal features, and one for generating the caption. Herein, the base image features used are the same $36 \times 2048$-dimensional features as in the ShowAttTell model. During the caption decoding, the computed attentional features are concatenated with the word embedding of the previous word in the caption, and input to the LSTM used for caption decoding.

\paragraph{Training and Optimization.} For each of the above systems, we report two sets of results. The first set of results was obtained by minimizing a cross-entropy loss (XE) for 30 epochs. The second set results was obtained by finetuning the cross-entropy-trained models using self-critical sequence training (SCST)~\cite{rennie2017self} for an additional 30 epochs with REINFORCE~\cite{sutton2000policy} over CIDEr~\cite{vedantam2015cider} metrics. During the finetuning, we used Adam~\cite{kingma2014adam} optimizer with an initial learning rate of $10^{-3}$ and a batch size of $128$.

\paragraph{Using Captioning Systems for the BISON Task.}
To use image-captioning algorithms to perform the BISON task, we compute the compatibility score of an image and caption by computing the log-likelihood of a caption given the associated image. A downside of this approach is that it tends to assign lower likelihood values to longer captions. To account for this, we normalize the compatibility score by dividing the log-likelihood by the caption length.

\subsection{Image Retrieval Systems}
We first describe the architecture of each of our image-retrieval systems, and then describe the process used to train these systems separately.

\paragraph{ConvNet+BoW.} We use two embedding networks to embed the image and text features into a joint embedding space. We use a one-hidden-layer multi-layer perceptron (MLP) to reduce the ($2048$-dimensional) averaged object proposal features (as described above) to $1024$ dimensions. For the text features, we compute average word embeddings of the caption and apply another MLP to embed the text features into the same space. As before, the word embeddings are initialized randomly. For both the image and the text MLP, the dimensionality of the hidden layer is $2048$ and a ReLU non-linearity is used. The system outputs the inner product between the image and caption features to measure the relevance of the image to the caption. 

\paragraph{ConvNet+Bi-GRU~\cite{Kiros2014UnifyingVE,faghri2018vse++}.} This baseline embeds image features using a MLP in the same fashion as \textit{ConvNet+BoW}. Different from ConvNet+BoW, a recurrent model (namely, a one-layer bidirectional GRU~\cite{chung2014empirical} (Bi-GRU) in our case) is used to map the word features into the same embedding space. In this Bi-GRU, the embedding is formed by (1) the average of last time step's output in the forward direction and (2) the first time step's output in the backward direction. The hidden dimensionality of the Bi-GRU is $1024$ because it is used to directly transform sequence of word embeddings to the joint embedding space.

\paragraph{OBJ+Bi-GRU.} Unlike \textit{ConvNet+Bi-GRU}, which takes the averaged object proposal features, this baseline uses a recurrent model as an aggregation function. Concretely, it uses a Bi-GRU to aggregate the $36 \times 2048$ dimensional visual feature into a feature vector of $2048$ dimensions. We input each object proposal feature into the Bi-GRU at each time step (36 steps in total; ordered by confidence of the object proposal) to obtain the $1024$ dimensional embedding. Similarly, a text Bi-GRU is used to embed the sequential word feature into the same embedding space. 

\paragraph{SCAN~\cite{lee2018stacked}.} We followed~\cite{lee2018stacked} to build this state-of-the-art image-text matching system with object proposal features and stacked cross-attention. We implement two variants of the SCAN system, namely, a variant with image-to-text attention (i2t) and a variant with text-to-image attention (t2i). We refer the reader to ~\cite{lee2018stacked} for complete details. 

\paragraph{Training and Optimization.} We follow~\cite{faghri2018vse++} to train the embedding networks for image and text, using a max-margin loss with hard negative mining. We use the Adam~\cite{kingma2014adam} optimizer for all methods. For \textit{ConvNet+BoW}, \textit{ConvNet+Bi-GRU} and \textit{OBJ+Bi-GRU}, we use a learning rate of $2\times10^{-4}$ with a batch size of $128$. Note that the max-margin objective we used is sensitive to the batch size. For \textit{SCAN}, we followed the hyper-parameter setting provided by ~\cite{lee2018stacked}, with a learning rate of $5\times10^{-4}$ and temperatures of $9$ for t2i variant and $6$ for i2t variant. We chose the average function as the aggregation function for \textit{SCAN}. Please refer to ~\cite{lee2018stacked} for details. 

\section{Effect of Visual Features}
\label{sec:experiment}

In this section, we provide additional results evaluating our image captioning and image retrieval systems using different sets of visual features.

\paragraph{Visual Features.}
We study the effect of varying the visual features by using two different ConvNets -- ResNet-IN~\cite{he2016deep} and ResNeXt-IG-3.5B~\cite{mahajan2018exploring}. \textit{ResNet-IN~\cite{he2016deep}} corresponds to the feature activations of the penultimate layer of a 101-layer deep residual network that was pre-trained on the ImageNet. \textit{ResNeXt-IG-3.5B~\cite{mahajan2018exploring}} corresponds to the feature activations of penultimate layer of a 101-layer ResNeXt model~\cite{xie2017aggregated} that was pre-trained on a weakly-supervised dataset of 3.5 billion images and corresponding hashtags. Using both networks, we first compute the $7 \times 7 \times 2048$ convolutional feature corresponding to a $224 \times 224$ input image. This feature is then reshaped to $49 \times 2048$ dimensions to replace the object proposal features we mentioned before.

\paragraph{Results.} Table~\ref{tab:captioning_eval} and Table~\ref{tab:retrieval_eval} report the BISON, image captioning, and image retrieval performances of systems using the aforementioned visual features. Comparing the results in these two tables with those in the main paper, we observe that using the ResNeXt-IG-3.5B feature~\cite{mahajan2018exploring} provides a boost in performance for all systems across all the evaluation measures.

\begin{table}[t]
	\tabcolsep 3pt
	\centering
	\resizebox{\linewidth}{!}{  
		\begin{tabular}{lccccc}\toprule
			Dataset $\rightarrow$& \multicolumn{4}{c}{COCO validation split} &  \OurDataset  \\
			Measure $\rightarrow$& \bf BLEU-4 & \bf CIDEr & \bf SPICE & \bf METEOR & \bf \GCSI  \\ \midrule
			\multicolumn{6}{l}{\bf Cross-entropy loss}  \\ \midrule
			
			\multicolumn{6}{l}{ \em with ResNet-IN Feature~\cite{he2016deep}} \\
			ShowTell~\cite{vinyals2015show} & 29.07 & 88.80 & 16.89 & 23.87 & 74.04 \\
			ShowAttTell~\cite{Xu2015ShowAA} & 32.09	& 99.30 & 18.47 & 25.24 & 79.32 \\ 
			UpDown~\cite{Anderson2017BottomUpAT}  & 31.87 & 99.82 & 18.67 & 25.34 & 81.39 \\
				
			\multicolumn{6}{l}{ \em with ResNeXt-IG-3.5B Feature~\cite{mahajan2018exploring}} \\
			ShowTell~\cite{vinyals2015show} & 33.08 & 102.41 & 18.84 & 25.97 & 81.85 \\
			ShowAttTell~\cite{Xu2015ShowAA} & 34.00	& 106.28 & 19.81 & 26.50 & 82.95 \\
			UpDown~\cite{Anderson2017BottomUpAT}  & 34.74 & 108.54 & 20.41 & 26.91 & 84.64 \\ 
			\midrule

			\multicolumn{6}{l}{\bf Self-critical sequence loss~\cite{rennie2017self}}  \\ \midrule
			
			\multicolumn{6}{l}{ \em with ResNet-IN Feature~\cite{he2016deep}} \\
			ShowTell~\cite{vinyals2015show} & 29.65 & 91.63 & 17.27 & 24.35 & 74.76 \\ 
			ShowAttTell~\cite{Xu2015ShowAA} & 32.06 & 100.24 & 18.71 & 25.36 & 79.91 \\ 
			UpDown~\cite{Anderson2017BottomUpAT}  & 32.63 & 102.35 & 19.00 & 25.83 & 81.84 \\

			\multicolumn{6}{l}{\em with ResNeXt-IG-3.5B Feature~\cite{mahajan2018exploring}} \\
			ShowTell~\cite{vinyals2015show}  & 33.37 & 104.18 & 19.30 & 26.39 & 82.13 \\
			ShowAttTell~\cite{Xu2015ShowAA}  & 34.19 & 107.22 & 19.99 & 26.64  & 83.39 \\
			UpDown~\cite{Anderson2017BottomUpAT}  & \bf 34.75 & \bf 109.51 & \bf 20.49 & \bf 27.15 & \bf 84.87 \\
			\midrule			
			
			Human~\cite{lsun-coco-workshop} & $21.7^{\star}$ & $85.4^{\star}$ & $19.8^{\star}$ & $25.2^{\star}$ & \bf 100.00 \\ \bottomrule
		\end{tabular}
	}
	\caption{\small \textbf{Performance of three image captioning systems} in terms of four captioning scores on the COCO validation set (left) and in terms of \GCSI accuracy on the COCO-BISON dataset (right). Human performances marked with $^\star$ were measured on the COCO test set. See text for details.}\label{tab:captioning_eval}
\end{table}

\begin{table}[h]
	\small
	\tabcolsep 3pt
	\centering
	\resizebox{\linewidth}{!}{ 
		\begin{tabular}{lccccc }\toprule
			Dataset $\rightarrow$ & \multicolumn{4}{c}{COCO-1K~\cite{karpathy2015deep}} & \OurDataset \\
			Task $\rightarrow$ & \multicolumn{2}{c}{\bf Image retrieval} & \multicolumn{2}{c}{\bf Caption retrieval} & \\
			Measure $\rightarrow$ &  \bf R@1 & \bf R@5 & \bf R@1 & \bf R@5 & \bf \GCSI \\ \midrule
			\multicolumn{6}{l}{\em with ResNet-IN Feature~\cite{he2016deep}} \\
			ConvNet+BoW & 40.12	& 74.79 & 51.36 & 81.36 & 77.96 \\
			ConvNet+Bi-GRU~\cite{Kiros2014UnifyingVE}  & 43.61 & 78.14 & 55.30 & 84.16 & 78.90 \\
			Obj+Bi-GRU  & 47.68 & 81.66 &  60.44 & 89.08 & 80.40 \\
			SCAN i2t~\cite{lee2018stacked} & 36.89	& 72.76 & 59.08	& 86.82 & 78.45 \\
			SCAN t2i~\cite{lee2018stacked} & 47.06 & 80.18 & 62.12 & 89.28 & 82.25 \\
			
			\multicolumn{6}{l}{\em with ResNeXt-IG-3.5B Feature~\cite{mahajan2018exploring}} \\
			ConvNet+BoW & 51.86 & 82.07 & 64.06 & 89.24 & 83.47 \\
			ConvNet+Bi-GRU~\cite{Kiros2014UnifyingVE}  & 53.85 & 84.75 & 65.38 & 90.18 & 84.23 \\
			Obj+Bi-GRU  & 51.90 & 83.98 & 63.78 & 89.18 & 82.24 \\
			SCAN i2t~\cite{lee2018stacked} & \bf 57.41 & \bf 86.86 & 68.62& \bf 92.72 & \bf 86.66 \\
			SCAN t2i~\cite{lee2018stacked} & 57.13 & 85.95 & \bf 69.28 & 92.38 & 86.14 \\ \bottomrule
		\end{tabular}
	}
	\caption{\small Quality of different systems for caption-based image retrieval (left) and image-based caption retrieval (right) in terms of recall at 1 and 5, compared to BISON scores. All models are evaluated on two sets of features --- ResNet-IN~\cite{he2016deep} and ResNext-IG-3.5B~\cite{mahajan2018exploring} features. Retrieval performances are reported on the COCO 1k test set of~\cite{karpathy2015deep} split. }
	\label{tab:retrieval_eval}
\end{table}

\begin{figure*}[tb!]
	\centering
	\begin{tabular}{cc}
		\includegraphics[width=0.485\linewidth]{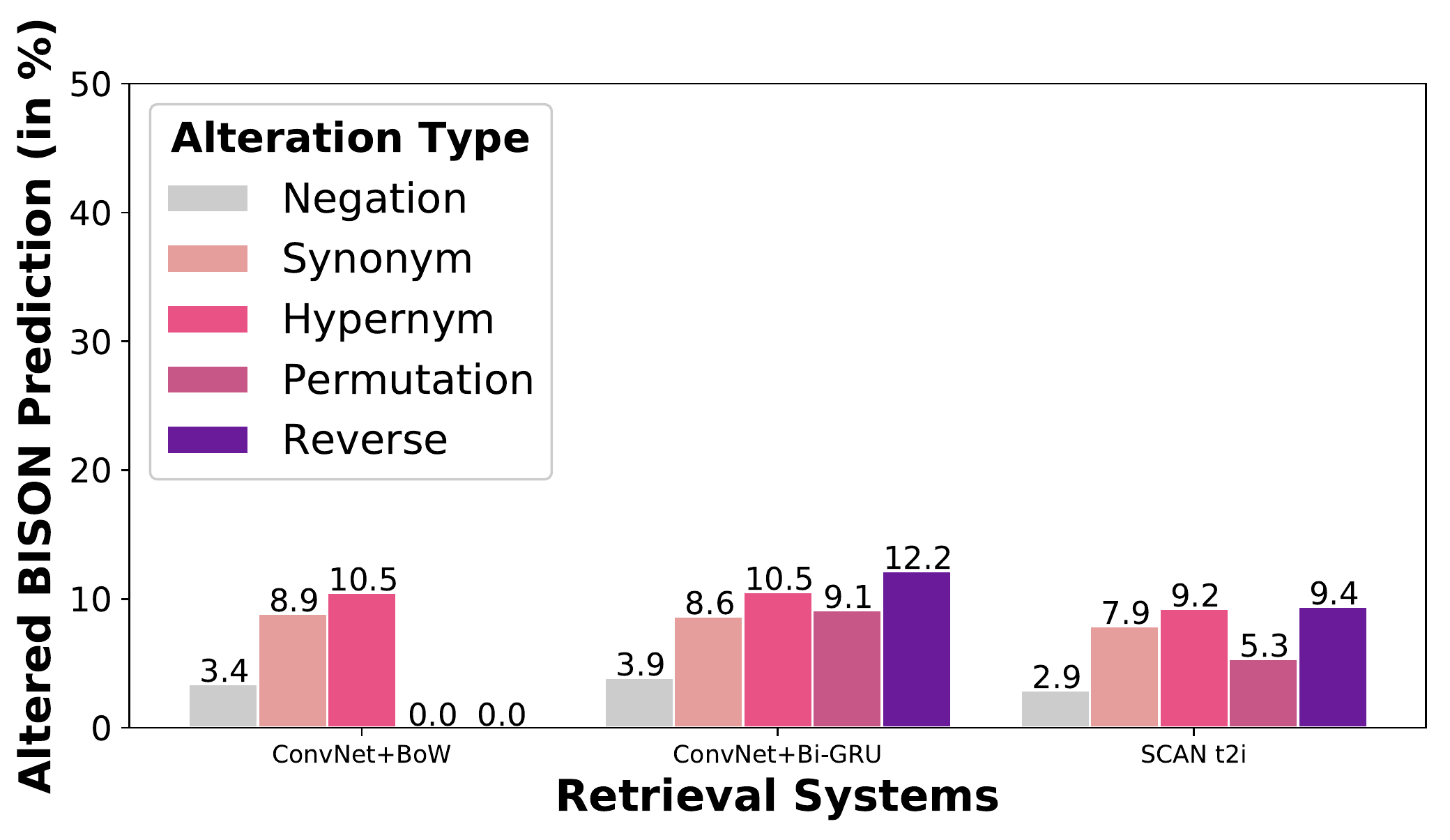} & 
		\includegraphics[width=0.485\linewidth]{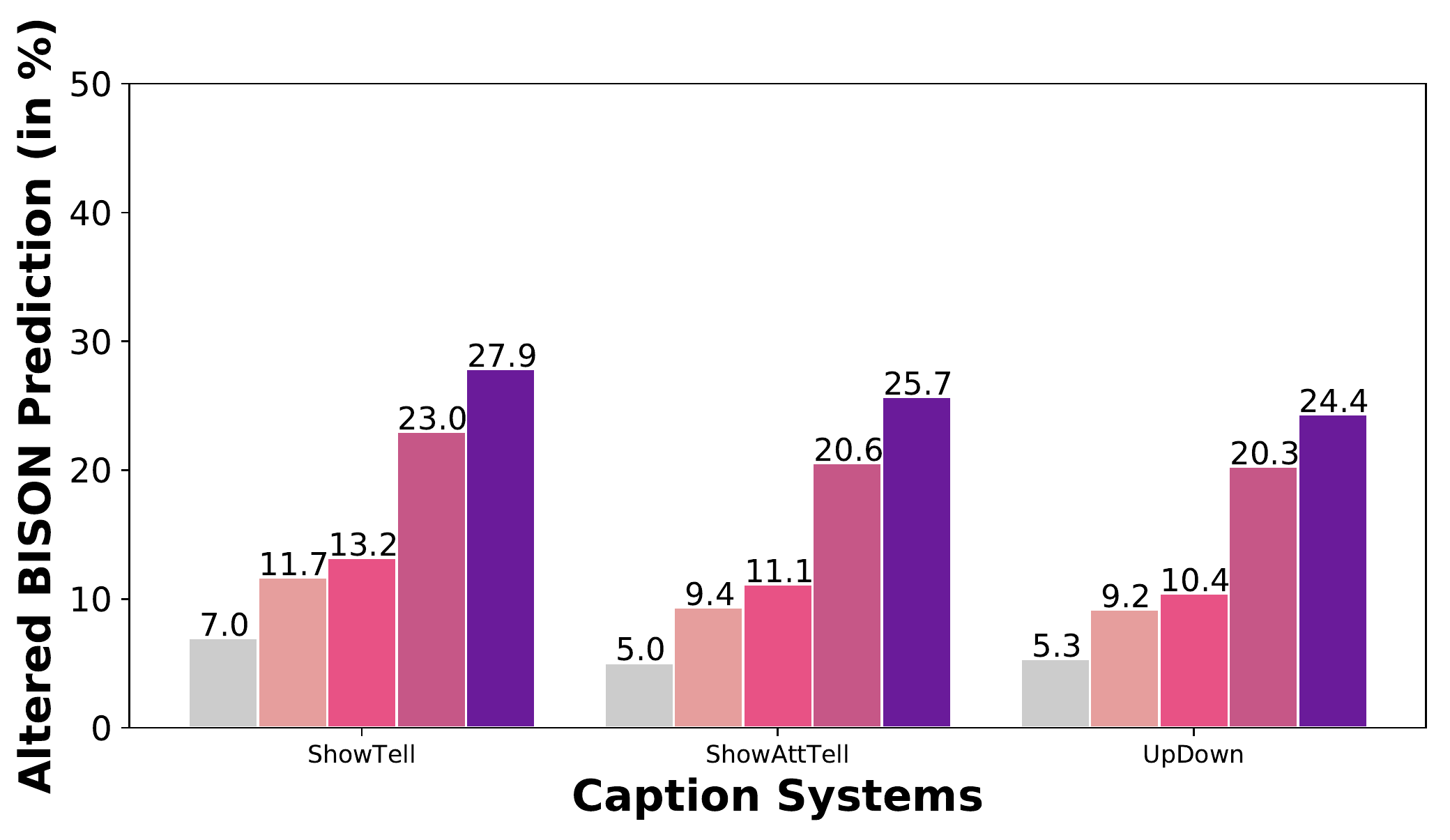}
	\end{tabular}
	\caption{\textbf{Changes in BISON prediction under automatic text-query perturbations.} 
		Percentage of predictions by various retrieval systems and captioning systems on COCO-BISON that are changed when words in the query are permuted or the query is negated \etc.}
	\label{more_query_alteration}
\end{figure*}

\section{Details on Analyzing Systems with BISON}
\label{sec:experiment}

In this section, we provide the complete details to the analysis we performed to evaluate systems with BISON (cf. Section 6 in the main paper)

\subsection{Details of Alteration on Text-query.}
We describe the details about the alteration we performed to the query text of a BISON triplet, which is unique to BISON evaluation.

\paragraph{Negation.} To a negate query text, we first perform a Part-of-Speech (POS) tagging to identify the word type of each word in the sentence with NLTK toolkit. Then we perform one step negation by adding not to the first verb word detected in the sentence. Heuristic regular expressions are applied to make sure that the resulting query text is a coherent English sentence. Next, this altered query text is used to evaluate a system on the original BISON pair of images. We show the proportion of changes in the systems' decisions and show them in Figure 9 in the main text. We observe that only a relatively small proportion of the system's decisions are changed, though the logical meaning of query is negated in most cases. A potential interpretation is that both caption systems and retrieval systems are not sensitive to the logical operation of negation.

\paragraph{Synonym \& Hypernym.} For the alteration of synonym and hypernym substitution, we first perform the Part-of-Speech (POS) tagging with NLTK toolkit. Then we ground all the nouns and verbs to the WordNet~\cite{pedersen2004wordnet} synsets  of the same word type and find the corresponding synonyms or hypernyms (we use the first lemma of the found synonyms/hypernyms). Next, during the alteration, we randomly sample one noun or verb and substitute it with one of its synonyms or hypernyms.

\paragraph{Permutation.} For the permutation alteration, we randomly permutate all the words within a sentence of the query text (Note that we exclude special word tokens such as ``\texttt{<bos>}'' or ``\texttt{<eos>}''). We found that captioning systems are much more sensitive to the word order whereas retrieval systems are less affected by this word permutation.

\paragraph{Reverse.} In addition to the random permutation, we further study the scenario in which the word order is exactly reversed (special word tokens are not affected). We observe that this causes more changes to the decisions of captioning systems.

\subsection{Analysis of Query Alteration on More Systems.}
We provide the same analysis to more captioning and retrieval systems and show the results as Figure~\ref{more_query_alteration}.

\begin{figure*}[b]
	\centering
	\begin{tabular}{cc}
		\includegraphics[width=0.7\textwidth]{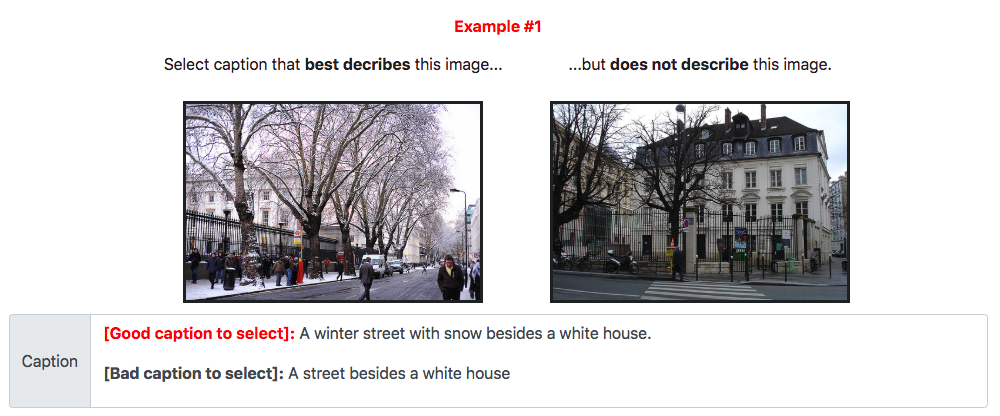} \\ 
		\includegraphics[width=0.7\textwidth]{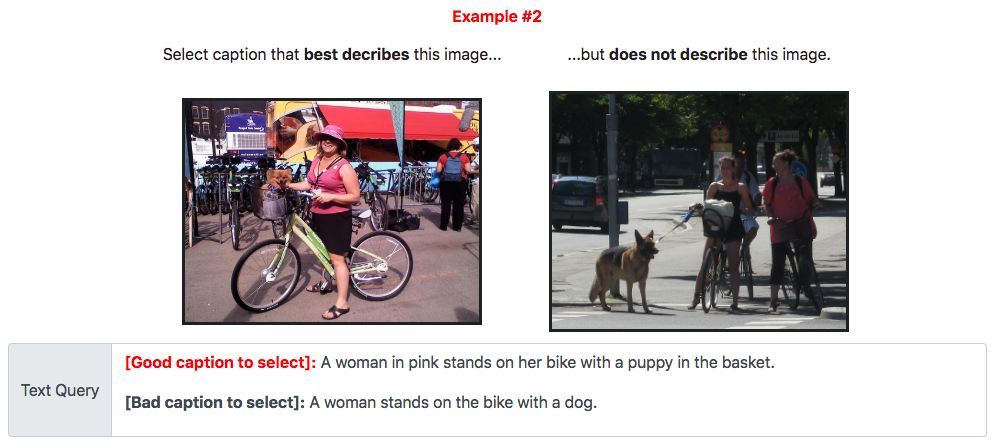} \\
		\includegraphics[width=0.7\textwidth]{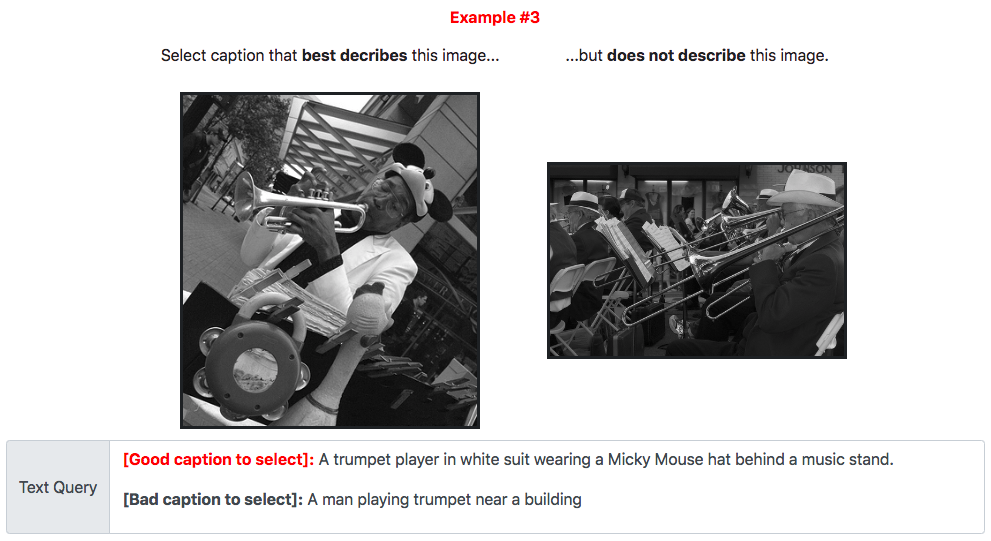}
	\end{tabular}
	\caption{Examples used in annotation stage 2 instruction \#3. }
	\label{identification_instruction_3}
\end{figure*}

\begin{figure*}[b]
	\centering
	\includegraphics[width=0.7\textwidth]{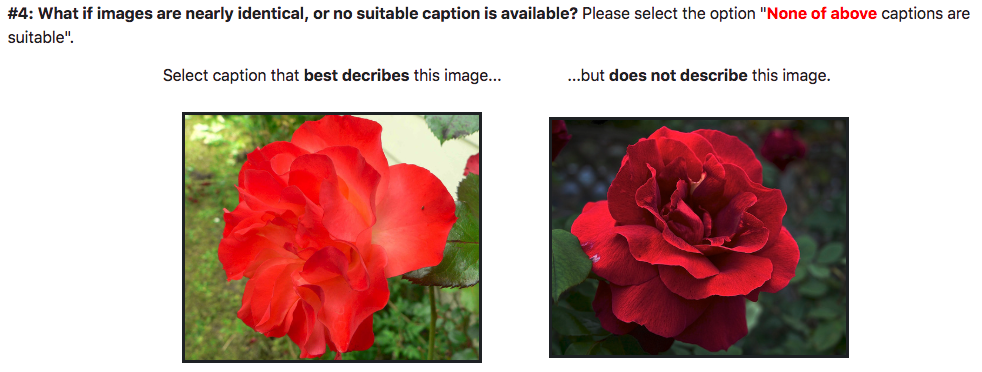}
	\caption{Examples used in annotation stage 2 instruction \#4. }
	\label{identification_instruction_4}
\end{figure*}

\begin{figure*}[b]
	\centering
	\includegraphics[width=0.7\textwidth]{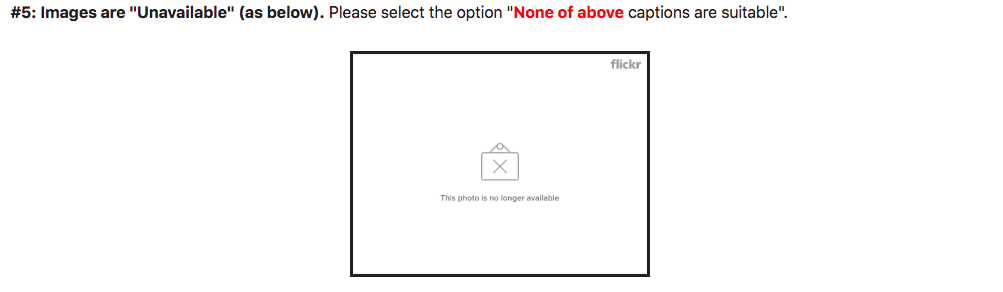}
	\caption{Examples used in stage 2 instruction \#5. }
	\label{identification_instruction_5}
\end{figure*}

\begin{figure*}[b]
	\centering
	\includegraphics[width=0.7\textwidth]{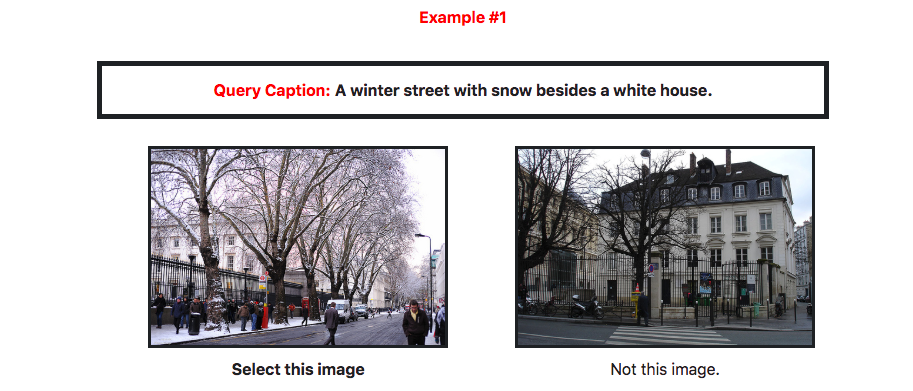}
	\caption{Examples used in stage 3 instruction \#3. }
	\label{verification_instruction_3}
\end{figure*}